\documentclass{article}

\usepackage{amsmath,amsfonts,amssymb,amsthm,bm}
\usepackage{enumitem}
\usepackage{microtype}
\usepackage{graphbox}
\usepackage{graphicx}
\usepackage{caption}
\usepackage{subcaption}
\usepackage{booktabs} 
\usepackage{amsfonts}       
\usepackage{nicefrac}       

\newcommand{\reals}{\mathbb{R}}

\newcommand{\erf}{\mathrm{erf}}
\newcommand{\EE}{\mathbb{E}\,}

\newcommand{\dd}{\operatorname{d}\!}

\usepackage{xcolor}
\definecolor{C0}{HTML}{1f77b4}
\definecolor{C1}{HTML}{ff7f0e}
\definecolor{C2}{HTML}{2ca02c}
\definecolor{C3}{HTML}{d62728}
\definecolor{C4}{HTML}{9467bd}
\definecolor{C5}{HTML}{8c564b}
\definecolor{C6}{HTML}{e377c2}
\definecolor{C7}{HTML}{7f7f7f}
\definecolor{C8}{HTML}{bcbd22}
\definecolor{C9}{HTML}{17becf}

\usepackage{hyperref}

\usepackage[accepted]{icml2021}

\icmltitlerunning{The Dynamics of Learning with Feedback Alignment}

\begin{document}

\twocolumn[
\icmltitle{Align, then memorise: \\the dynamics of learning with feedback alignment}



\icmlsetsymbol{equal}{*}

\begin{icmlauthorlist}
\icmlauthor{Maria Refinetti}{equal,ens,ide}
\icmlauthor{St\'ephane d'Ascoli}{equal,ens,fair}
\icmlauthor{Ruben Ohana}{ens,lighton}
\icmlauthor{Sebastian Goldt}{sissa}
\end{icmlauthorlist}

\icmlaffiliation{ens}{Department of Physics, Ecole Normale Sup\'erieure, Paris,
  France}
\icmlaffiliation{fair}{Facebook AI Research, Paris, France}
\icmlaffiliation{lighton}{LightOn, Paris, France}
\icmlaffiliation{sissa}{International School of Advanced Studies (SISSA), Trieste, Italy}
\icmlaffiliation{ide}{IdePHICS laboratory, EPFL}

\icmlcorrespondingauthor{Sebastian Goldt}{sgoldt@sissa.it}

\icmlkeywords{Machine Learning, ICML}

\vskip 0.3in
]



\printAffiliationsAndNotice{\icmlEqualContribution} 

\begin{abstract}
  Direct Feedback Alignment (DFA) is emerging as an efficient and biologically
  plausible alternative to backpropagation for training deep neural networks.
  Despite 
  relying on random feedback weights for the backward pass, DFA successfully trains
  state-of-the-art models such as Transformers.
  On the other hand, it notoriously fails to train convolutional networks.
  An understanding of the inner workings of DFA to explain these diverging
  results remains elusive.
  Here, we propose a theory of feedback alignment algorithms.
  We first show that learning in shallow networks proceeds in two steps: an
  \emph{alignment} phase, where the model adapts its weights to align the
  approximate gradient with the true gradient of the loss function, is followed by a
  \emph{memorisation} phase, where the model focuses on fitting the data.
  This two-step process has a \emph{degeneracy breaking} effect: out of all the
  low-loss solutions in the landscape, a network trained with DFA naturally converges to the
  solution which maximises gradient alignment.
  We also identify a key quantity underlying alignment in deep linear networks:
  the conditioning of the \emph{alignment matrices}.  The latter enables a
  detailed understanding of the impact of data structure on alignment, and
  suggests a simple explanation for the well-known failure of DFA to train
  convolutional neural networks.
  Numerical experiments on MNIST and CIFAR10 clearly demonstrate degeneracy
  breaking in deep non-linear networks and show that the align-then-memorize
  process occurs sequentially from the bottom layers of the network to the top.
\end{abstract}

\vspace*{1em}

\section*{Introduction} 
\begin{figure}[t!]
  \centering
  \includegraphics[width=.5\textwidth]{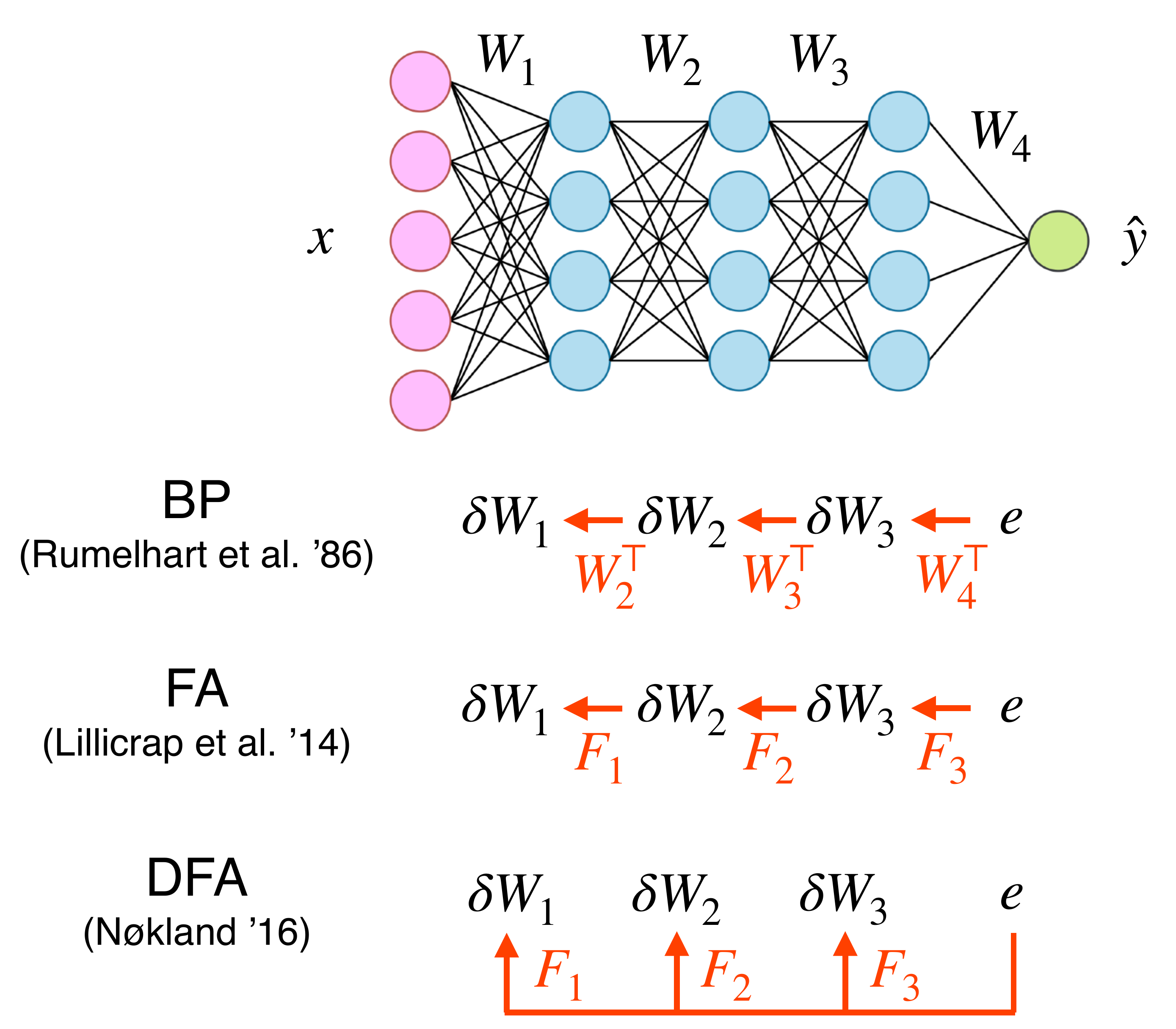}
  \caption{\label{fig:algorithms} \textbf{Three approaches to the credit assignment problem in deep
      neural networks.} In \emph{back-propagation} (BP), the weight updates
    $\delta W_l$ are computed sequentially by transmitting the error $e$ from
    layer to layer using the transpose of the network's weights
    $W_l^\top$. In \emph{feedback alignment} (FA)~\cite{lillicrap2016random},
    $W_l^\top$ are replaced by fixed random feedback matrices $F_l$. In
    \emph{direct feedback alignment} (DFA)~\cite{noekland2016direct}, the error is
    directly injected to each layer using random feedback matrices $F_l$,
    enabling parallelized training.}
\end{figure}

Training a deep neural network on a supervised learning task requires solving
the credit assignment problem: how should weights deep in the network be
changed, given only the output of the network and the target label of the input?
Today, almost all networks from computer vision to natural language processing
solve this problem using variants of the back-propagation algorithm (BP)
popularised several decades ago by~\citet{rumelhart1986learning}. For
concreteness, we illustrate BP using a fully-connected deep network of depth
$L$ with weights $W_l$ in the~$l$th layer
. Given an input $x\equiv h_0$, the output~$\hat y$ of the network is computed
sequentially as $\hat y = f_y (a_L)$, with $a_{l} = W_l h_{l-1}$ and
$h_l = g(a_l)$, where $g$ is a pointwise
non-linearity. 
For regression, the loss function $J$ is the mean-square error
and $f_y$ is the identity.  Given the error
$e \equiv \nicefrac{\partial J}{\partial a_{L}}=\hat{y}-y$ of the network on an
input $x$, the update of the last layer of weights reads
\begin{equation}
  \delta W_L = -\eta e h_{L-1}^\top
\end{equation}
for a learning rate $\eta$. The updates of the layers below are given by
$\delta W_{l}=-\eta \delta a_{l} h_{l-1}^{T}$, with factors $\delta a_l $
defined sequentially as
\begin{align}
  \label{eq:bp-update}
  \delta a_{l}^{\mathrm{BP}}=\nicefrac{\partial J}{\partial a_l} = \left(W_{l+1}^{T} \delta a_{l+1}\right) \odot g^{\prime}\left(a_{l}\right),
\end{align}
with $\odot$ denoting the Hadamard product.  BP thus solves the credit
assignment problem for deeper layers of the network by using the transpose of
the network's weight matrices to transmit the error signal across the network
from one layer to the next, see Fig.~\ref{fig:algorithms}.

Despite its popularity and practical success, BP suffers from several
limitations. First, it relies on symmetric weights for the forward and backward
pass, which makes it a biologically implausible learning algorithm~\cite{grossberg1987competitive, crick1989recent}. Second, BP updates layers sequentially during the backward pass, preventing an efficient parallelisation of training, which becomes ever
more important as state-of-the-art networks grow larger and deeper.

In light of these shortcomings, algorithms which only approximate the gradient
of the loss are attracting increasing interest.  \citet{lillicrap2016random}
demonstrated that neural networks can be trained successfully even if the
transpose of the network weights $W_l^\top$ are replaced by \emph{random}
feedback connections $F_l$ in the backward pass, an algorithm they called
``feedback alignment''~(FA):
\begin{align}
  \delta a_{l}^{\text{FA}}=\left(F_{l} \delta a_{l+1}\right) \odot g^{\prime}\left(a_{l}\right).
\label{eq:fa-update}
\end{align}
In this way, they dispense with the need of biologically unrealistic symmetric
forward and backward weights~\cite{grossberg1987competitive, crick1989recent}.
The ``direct feedback alignment'' (DFA) algorithm of~\citet{noekland2016direct}
takes this idea one step further by propagating the error directly from the
output layer to each hidden layer of the network through random feedback
connections $F_l$:
\begin{equation}
  \label{eq:dfa-update}
  \delta a_{l}^{\text{DFA}}=\left(F_{l} e\right) \odot g^{\prime}\left(a_{l}\right).
\end{equation}
DFA thus allows updating different layers in parallel. Fig.~\ref{fig:algorithms} shows the information flow of all three algorithms.

While it was initially unclear whether DFA could scale to challenging datasets
and complex architectures~\cite{gilmer2017explaining,bartunov2018assessing},
recently \citet{launay2020direct} obtained performances comparable to fine-tuned
BP when using DFA to train a number of state-of-the-art architectures on
problems ranging from neural view synthesis to natural language processing. Yet,
feedback alignment notoriously fails to train convolutional
networks~\cite{bartunov2018assessing, moskovitz2018feedback,
  launay2019principled, han2019direct}. These varied results underline the need for a theoretical understanding of how
and when feedback alignment works. 

\paragraph{Related Work} \citet{lillicrap2016random} gave a first theoretical characterisation of
feedback alignment by arguing that for two-layer linear networks, FA works because the transpose of the second
layer of weights $W_2$ tends to align with the random feedback matrix $F_1$
during training. This \emph{weight alignment} (WA) leads the weight updates of
FA to align with those of BP, leading to \emph{gradient alignment} (GA) and thus to
successful learning.
\citet{frenkel2019learning} extended this analysis to the
deep linear case for a variant of DFA called ``Direct Random Target Projection''
(DRTP), 
under the restrictive assumption of training on a single data
point. \citet{noekland2016direct} also introduced a layerwise alignment criterion to
describe DFA in the deep nonlinear setup, under the assumption of constant
update directions for each data point.

\paragraph{Contributions}

\begin{enumerate}[leftmargin=*]
\item We give an analytical description of DFA dynamics in shallow non-linear
  networks, building on seminal work analysing BP in the limit of infinitely
  many training samples~\cite{saad1995a, saad1995b, biehl1995}.
\item We show that in this setup, DFA proceeds in two steps: an alignment
  phase, where the forward weights adapt to the feedback weights to improve the approximation of the gradient, is followed
  by a memorisation phase, where the network sacrifices some alignment to
  minimise the loss. Out of
  the same-loss-solutions in the landscape, DFA converges to the one that
  maximises gradient alignment, an effect we term
  ``degeneracy breaking''.
\item We then focus on the alignment phase in the setup of deep linear networks,
  and uncover a key quantity underlying GA: the conditioning of the alignment
  matrices. Our framework allows us to analyse the impact of data structure on
  DFA, and suggests an explanation for the failure of DFA to train convolutional
  layers.
\item We complement our theoretical results with experiments that demonstrate
  the occurence of (i) the Align-then-Memorise phases of learning, (ii)
  degeneracy breaking and (iii) layer-wise alignment in deep neural networks
  trained on standard vision
  datasets.  
\end{enumerate}

\paragraph{Reproducibility} We host all the code to reproduce our experiments
online at~\url{https://github.com/sdascoli/dfa-dynamics}.

\begin{figure*}[t!]
  \begin{subfigure}[b]{.33\textwidth}
    \includegraphics[width=\linewidth]{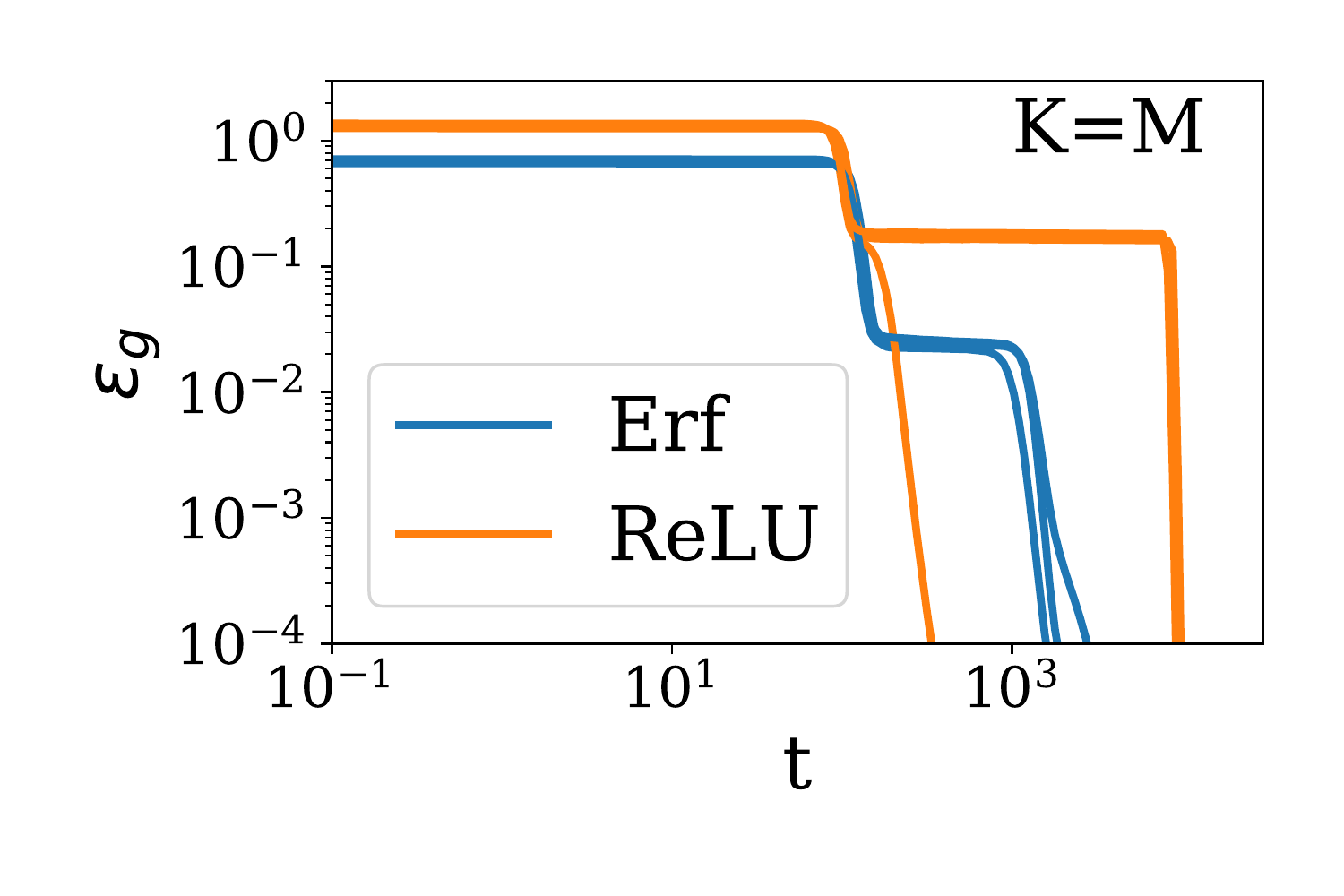}%
    \vspace*{-.5cm}
    \caption{BP, matched}
    \end{subfigure} 
    \begin{subfigure}[b]{.33\textwidth}
    \includegraphics[width=\linewidth]{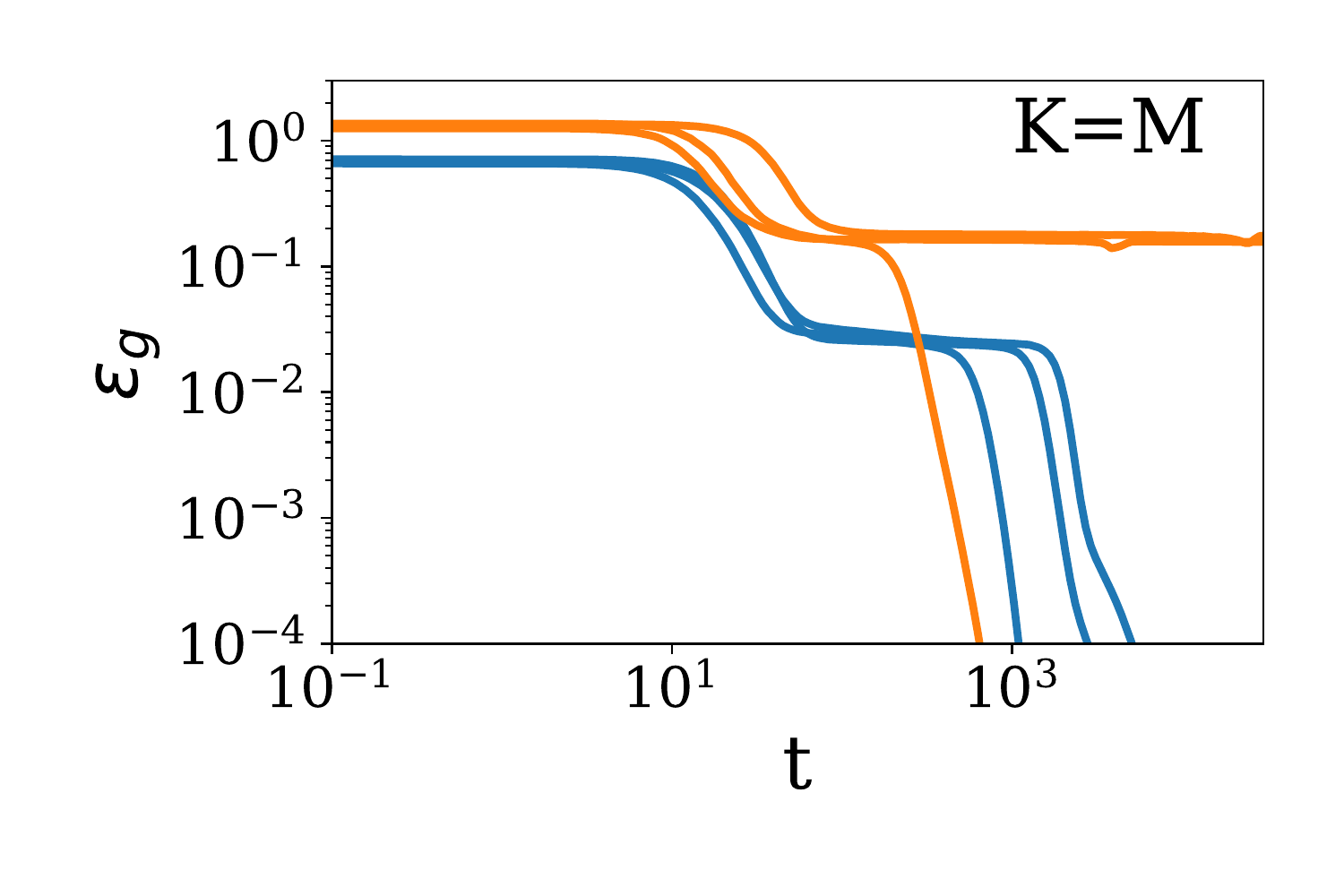}%
    \vspace*{-.5cm}
    \caption{DFA, matched}
    \end{subfigure} 
    \begin{subfigure}[b]{.33\textwidth}
    \includegraphics[width=\linewidth]{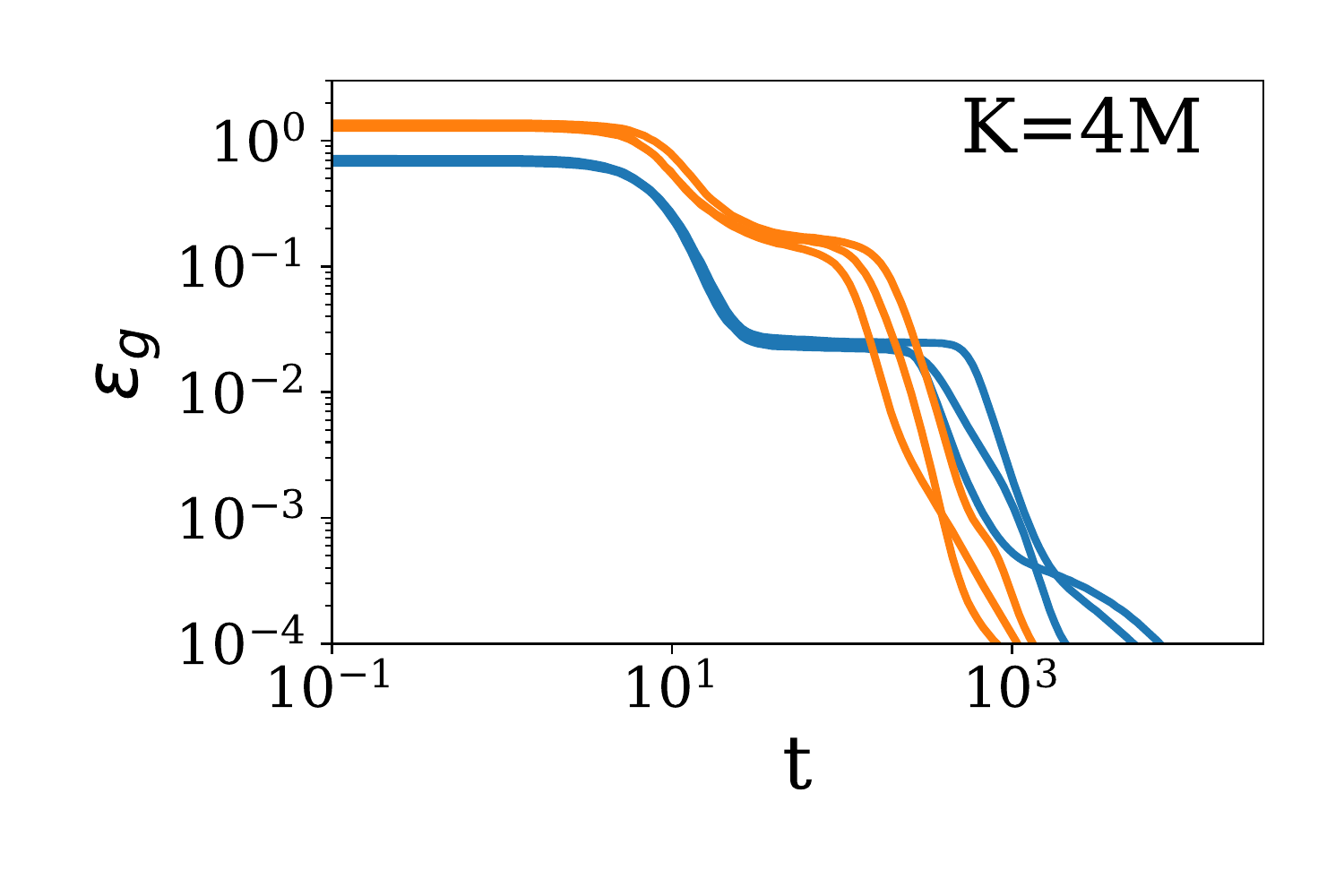}%
    \vspace*{-.5cm}
    \caption{DFA, overparametrized}
    \end{subfigure} 
	\caption{ \textbf{Learning dynamics of back-propagation and feedback
        alignment for \textcolor{C0}{sigmoidal} and \textcolor{C1}{ReLU} neural
        networks learning a target function.} Each plot shows three runs from
      different initial conditions for every setting, where a shallow neural
      network with $K$ hidden nodes tries to learn a teacher network with $M$
      hidden nodes. \emph{(a)} All networks trained using BP in the matched
      case $K=M$ achieve perfect test error. \emph{(b)} Sigmoidal networks
      achieve perfect test error with DFA, but the algorithm fails in some
      instances to train ReLU networks ($K=M$) \emph{(c)} In the
      over-parametrised case ($K>M$), both sigmoidal and ReLU networks achieve
      perfect generalisation when trained with DFA. \emph{Parameters:}
      $N=500, L=2, M=2, \eta=0.1, \sigma_0 = 10^{-2}$.}
     \label{fig:dynamics}
\end{figure*}
\section{A two-phase learning process}
\label{sec:db}

We begin with an exact description of DFA dynamics in shallow non-linear
networks. Here we consider a high-dimensional scalar regression task where the inputs
$x \in \reals^{N}$ are sampled i.i.d.\ from the standard normal distribution. We
focus on the classic \emph{teacher-student} setup, where the labels~$y\in\reals$
are given by the outputs of a ``teacher'' network with random
weights~\cite{gardner1989, seung1992, watkin1993, engel2001statistical, zdeborova2016}. In
this section, we let the input dimension $N\to\infty$, while both teacher and
student are two-layer networks with $K,M\sim O(1)$ hidden nodes.

We consider sigmoidal, $g(x)=\erf\left( \nicefrac{x}{\sqrt{2}} \right)$, and
ReLU activation functions, $g(x)=\max(0, x)$. We asses the student's performance on the task through its
the \emph{generalisation error}, or test error:
\begin{equation}
  \label{eq:eg}
  \epsilon_g(\theta, \tilde \theta) \equiv \frac{1}{2}\EE{\left[ \hat y - y \right]}^2 \equiv \frac{1}{2}\EE\left[{e^2}\right],
\end{equation}
where the expectation $\EE$ is taken over the inputs for a given teacher and
student networks with parameters
$\tilde \theta=(M, \tilde W_1, \tilde W_2, g)$ and
$\theta=(K, W_1, W_2, g)$. Learning a target function such as the teacher is
a widely studied setup in the theory of neural networks~\cite{zhong2017recovery,
  advani2020high, tian2017analytical, du2018gradient,
  soltanolkotabi2018theoretical, aubin2018committee, saxe2018information,
  baity-jesi2018, goldt2019dynamics, ghorbani2019limitations,
  yoshida2019datadependence, bahri2020statistical, gabrie2020meanfield}.

In this shallow setup, FA and DFA are equivalent, and only involve one feedback
matrix, $F_1\in\mathbb R^{K}$ which back-propagates the error signal $e$ to the first
layer weights $W_1$. The updates of the second layer of weights $W_2$ are the
same as for BP.

\paragraph{Performance of BP vs. DFA} We show the evolution of the test
error~\eqref{eq:eg} of sigmoidal and ReLU students trained via vanilla BP in the
``matched'' case $K=M$ in Fig.~\ref{fig:dynamics}~a, for three random choices of
the initial weights with standard deviation $\sigma_0=10^{-2}$. In all cases,
learning proceeds in three phases: an initial exponential decay; a phase where
the error stays constant, the ``plateau''~\cite{saad1995a, engel2001statistical,
  yoshida2019datadependence}; and finally another exponential decay towards zero
test error.

Sigmoidal students trained by DFA always achieve perfect generalisation when started from
different initial weights with a different feedback vector each time (blue in
Fig.~\ref{fig:dynamics} b) raising a first question: if the
student has to align its second-layer weights with the random feedback vector in
order to retrieve the BP gradient~\cite{lillicrap2016random}, i.e. $W_2\propto F_1$, how can it recover the teacher weights perfectly, i.e. $W_2=\tilde W_2$?

For ReLU networks, over-parametrisation is key to the consistent success of DFA:
while some students with $K=M$ fail to reach zero test error (orange in
Fig.~\ref{fig:dynamics} b), almost every ReLU student having more parameters
than her teacher learns perfectly ($K=4M$ in Fig.~\ref{fig:dynamics} c). A second question follows: how does over-parameterisation help ReLU students
achieve zero test error?

\paragraph{An analytical theory for DFA dynamics} To answer these two questions,
we study the dynamics of DFA in the limit of infinite training data where a
previously unseen sample $(x, y)$ is used to compute the DFA weight
updates~(\ref{eq:dfa-update}) at every step. This ``online learning'' or
``one-shot/single-pass'' limit of SGD has been widely studied in recent and
classical works on vanilla BP~\cite{kinzel1990, biehl1995, saad1995a, saad1995b,
  saad2009line, zhong2017recovery, brutzkus2017globally, mei2018, rotskoff2018,
  chizat2018, sirignano2018}.
\begin{figure*}[t!]
\centering
    \begin{subfigure}[b]{.33\textwidth}
    \includegraphics[width=\linewidth]{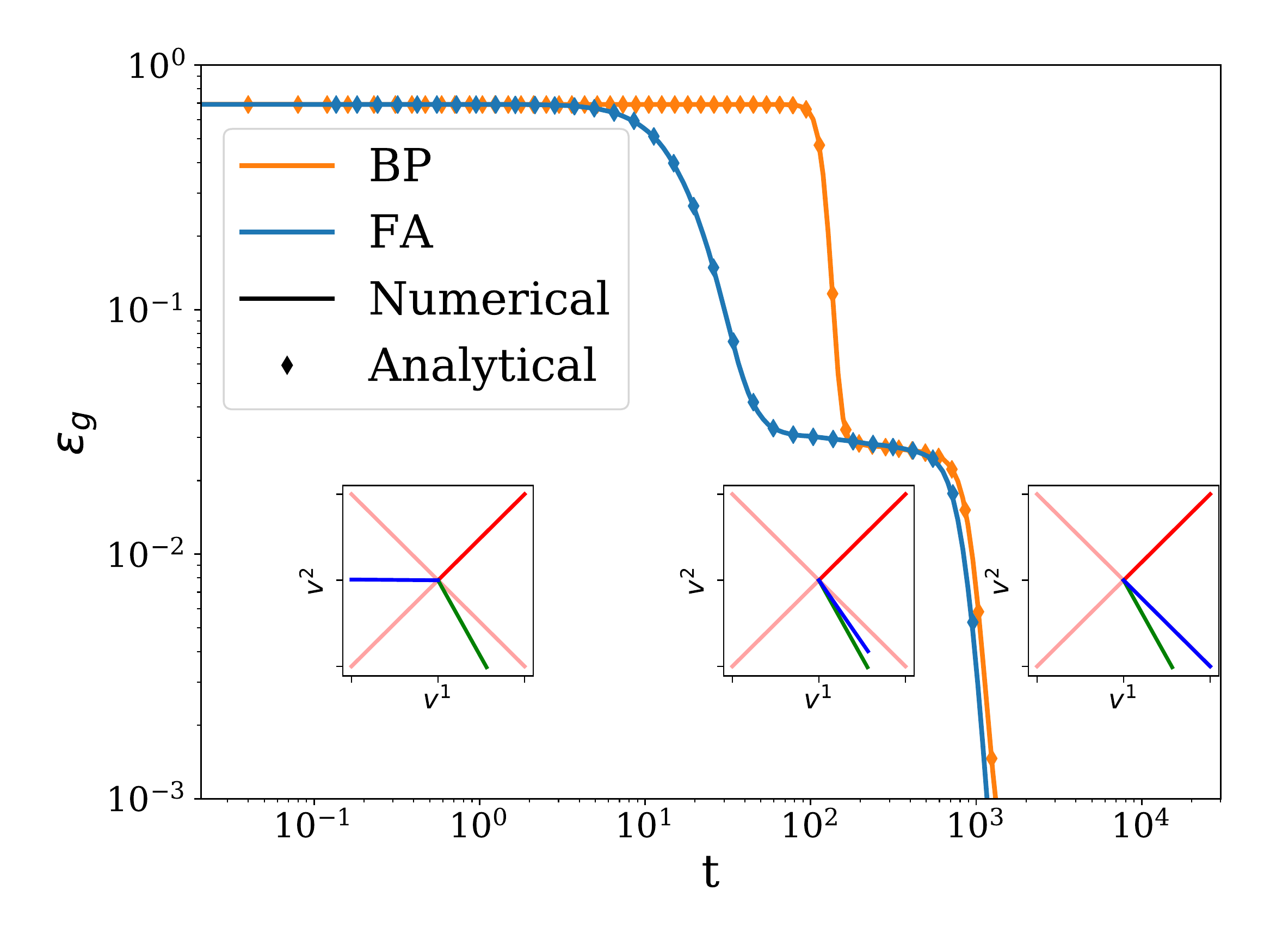}%
    \caption{Generalization dynamics}
    \end{subfigure} 
    \begin{subfigure}[b]{.33\textwidth}
    \includegraphics[width=\linewidth]{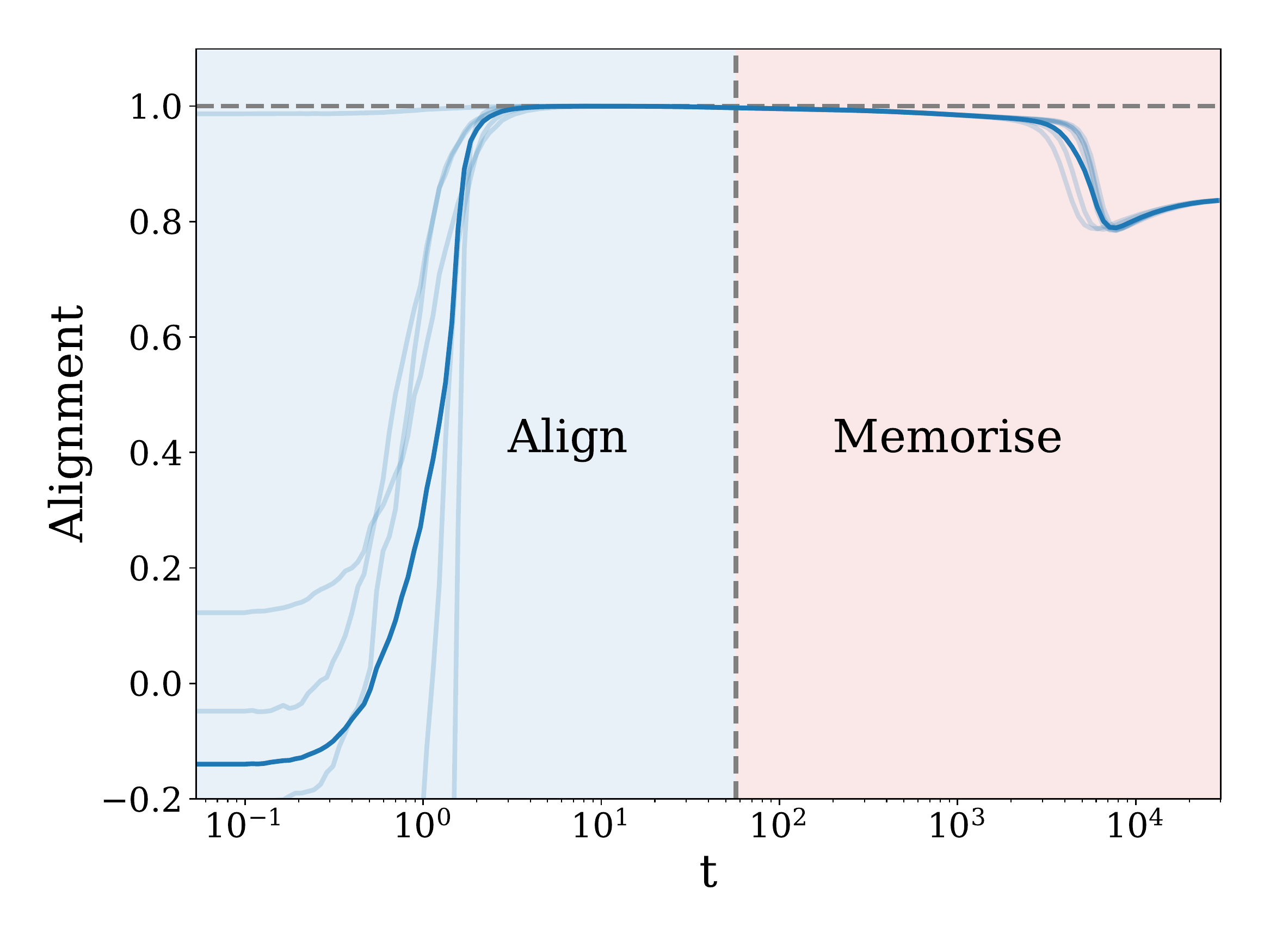}
    \caption{Alignment dynamics}
    \end{subfigure} 
    \hfill
    \begin{subfigure}[b]{.30\textwidth}
    \includegraphics[width=\linewidth]{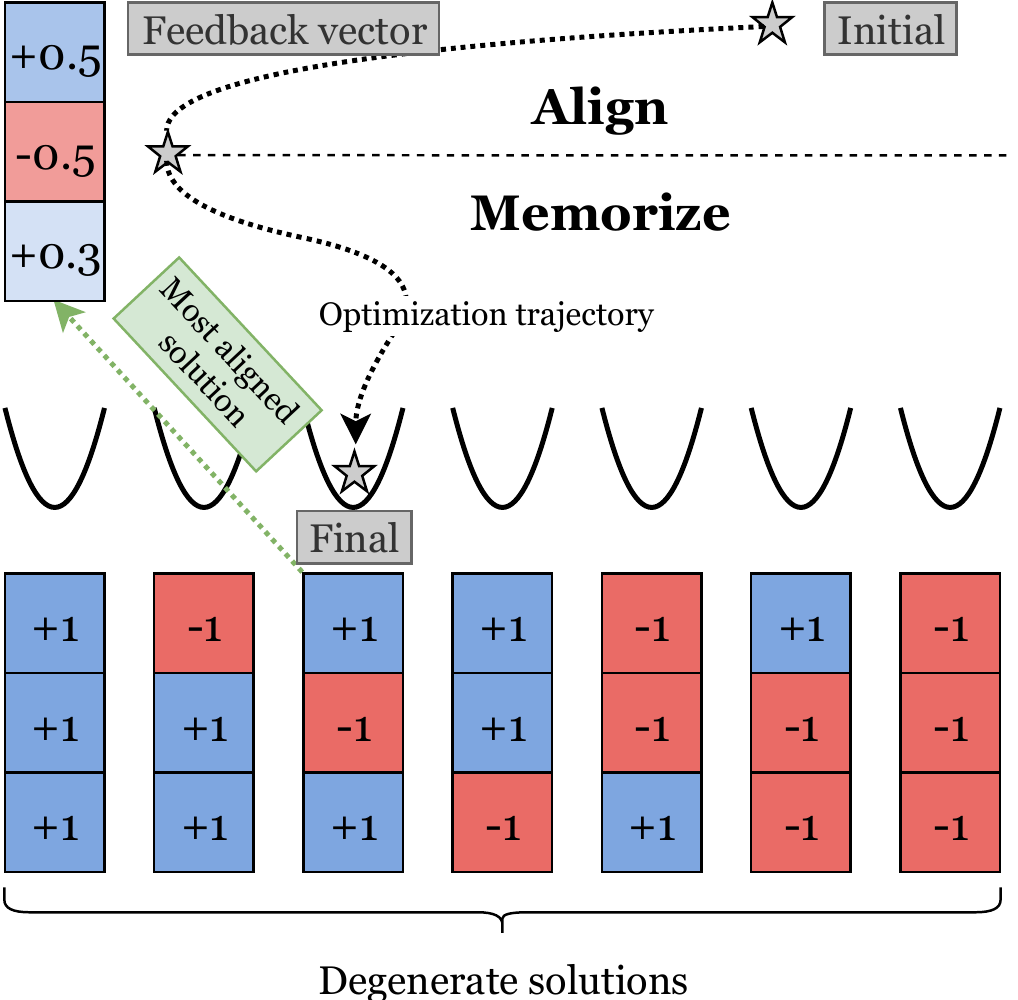}
    \caption{Degeneracy breaking}
    \end{subfigure} 

	\caption{ \label{fig:degeneracy_breaking} \emph{(a)} \textbf{Theory gives
        exact prediction for the learning dynamics.} We plot learning curves for
      BP and DFA obtained from (i) a single simulation (solid lines), (ii)
      integration of the ODEs for BP dynamics~\cite{biehl1995, saad1995a}
      (\textcolor{C1}{orange} dots), (iii) integration of the ODEs for DFA
      derived here (\textcolor{C0}{blue} dots). \emph{Insets}: Teacher
      second-layer weights (red) as well as the degenerate solutions (light red)
      together with the feedback vector $F_1$ (green) and the student second-layer
      weights $v$ (blue) at three different times during training with
      DFA. \emph{Parameters:} $N=500, K=M=2, \eta=0.1, \sigma_0 = 10^{-2}$.\newline
      \emph{(b)} \textbf{Align-then-Memorise process.}  Alignment (cosine
      similarity) between the student's second layer weights and the feedback
      vector. In the align phase, the alignment increases, and reaches its
      maximal value when the test loss reaches the plateau. Then it decreases in
      the memorization phase, as the student recovers the teacher weights.\newline
      \emph{(c)} \textbf{The degeneracy breaking mechanism.} There are multiple
        degenerate global minima in the optimisation landscape: they are related
        through a discrete symmetry transformation of the weights that leaves
        the student's output unchanged. DFA chooses the solution which maximises
        the alignment with the feedback vector.}
\end{figure*}

We work in the regime where the input dimension $N\to\infty$, while
$M$ and $K$ are finite. The test error~\eqref{eq:eg}, i.e.~a function of the student and teacher parameters involving
a high-dimensional average over inputs, can be simply expressed in terms of a
\emph{finite} number of ``order parameters''
$Q=(Q^{kl}), R=(R^{km}), T=(T^{mn})$,
\begin{equation}
  \label{eq:eg_order}
  \lim_{N\to\infty} \epsilon_g(\theta, \tilde \theta)
  = \epsilon_g(Q, R, T, W_2, \tilde W_2)
\end{equation}
where
\begin{equation}
    \label{eq:OP}
    Q^{kl}\!=\!\frac{W_1^k W_1^l}{N}, \text{ }
    R^{km}\!=\!\frac{W_1^k\tilde W_1^m}{N}, \text{ }
    T^{mn}\!=\!\frac{\tilde W_1^m\tilde W_1^n}{N}
\end{equation}
as well as second layer weights $\tilde W_2^{m}$ and $W_2^k$~\cite{saad1995a,
  saad1995b, biehl1995, engel2001statistical}. 
Intuitively, $R^{km}$ quantifies the similarity between the
weights of the student's $k$th hidden unit and the teacher's $m$th hidden unit. The self-overlap of the $k$th and $l$th student nodes is given by
$Q^{kl}$, and likewise $T^{mn}$ gives the (static) self-overlap of teacher
nodes. In seminal work, \citet{saad1995a} and \citet{biehl1995} obtained a
closed set of ordinary differential equations (ODEs) for the time evolution of
the order parameters $Q$ and $R$. Our first main contribution is to extend their
approach to the DFA setup (see SM~\ref{sec:EOM} for the details), obtaining a
set of ODEs~\eqref{eq:eom} that predicts the test error of a student trained
using DFA~(\ref{eq:dfa-update}) at all times. 
The accuracy of the predictions from the ODEs is demonstrated in Fig.~\ref{fig:degeneracy_breaking}~a, where the comparison between a single simulation of training a two-layer net with BP (orange) and DFA (blue) and theoretical predictions yield perfect agreement.

\subsection{Sigmoidal networks learn through ``degeneracy breaking''}

The test loss of a sigmoidal student trained on a teacher with the same number
of neurons as herself ($K=M$) contains several global minima, which all
correspond to fixed points of the ODEs~\eqref{eq:eom}. Among these is a student with exactly the same weights as her teacher. The
symmetry $\erf(z)=-\erf(-z)$ induces a student with weights $\{\tilde W_1, \tilde W_2\}$ to have the same test error as a sigmoidal student with weights $\{-\tilde W_1,-\tilde W_2\}$. Thus, as illustrated in Fig.~\ref{fig:degeneracy_breaking}~c, the problem of learning a teacher has various degenerate solutions. A student trained with vanilla BP converges to any one of these solutions, depending on the initial conditions. 

\paragraph{Alignment phase} A student trained using DFA has to fulfil
the same objective (zero test error), with an additional constraint: her
second-layer weights $W_2$ need to align with the feedback vector~$F_1$ to ensure the first-layer weights are updated in the direction that minimises the test error. 
And indeed, an analysis of the ODEs
(cf.~Sec.~\ref{sec:early-stages}) reveals that in the early phase of training,
$\dot{W_2} \sim F$ and so $W_2$ grows in the direction of the feedback vector
$F_1$ resulting in an increasing overlap between $W_2$ and $F_1$. 
In this \emph{alignment phase} of learning, shown in  Fig.~\ref{fig:degeneracy_breaking}~b, $W_2$ becomes perfectly aligned with $F_1$.
DFA has perfectly recovered the weight updates for $W_1$ of BP, but the second layer has lost its expressivity (it is simply aligned to the random feedback
vector). 

\paragraph{Memorisation phase} The expressivity of the student is restored in the \emph{memorisation}
phase of learning, where the second layer weights move away from~$F_1$ and towards
the global miminum of the test error that maintains the highest overlap with the
feedback vector. 
In other words, students solve this constrained optimisation problem by consistently
converging to the global minimum of the test loss that simultaneously maximises
the overlap between $W_2$ and $F_1$, and thus between the DFA gradient and the BP gradient. For DFA, the global minima of the test loss
are not equivalent, this ``degeneracy breaking'' is illustrated in Fig.~\ref{fig:degeneracy_breaking}~c.

\subsection{Degeneracy breaking requires over-parametrisation for ReLU networks}
    
\begin{figure}[t!]
  \centering
  \includegraphics[width=\columnwidth]{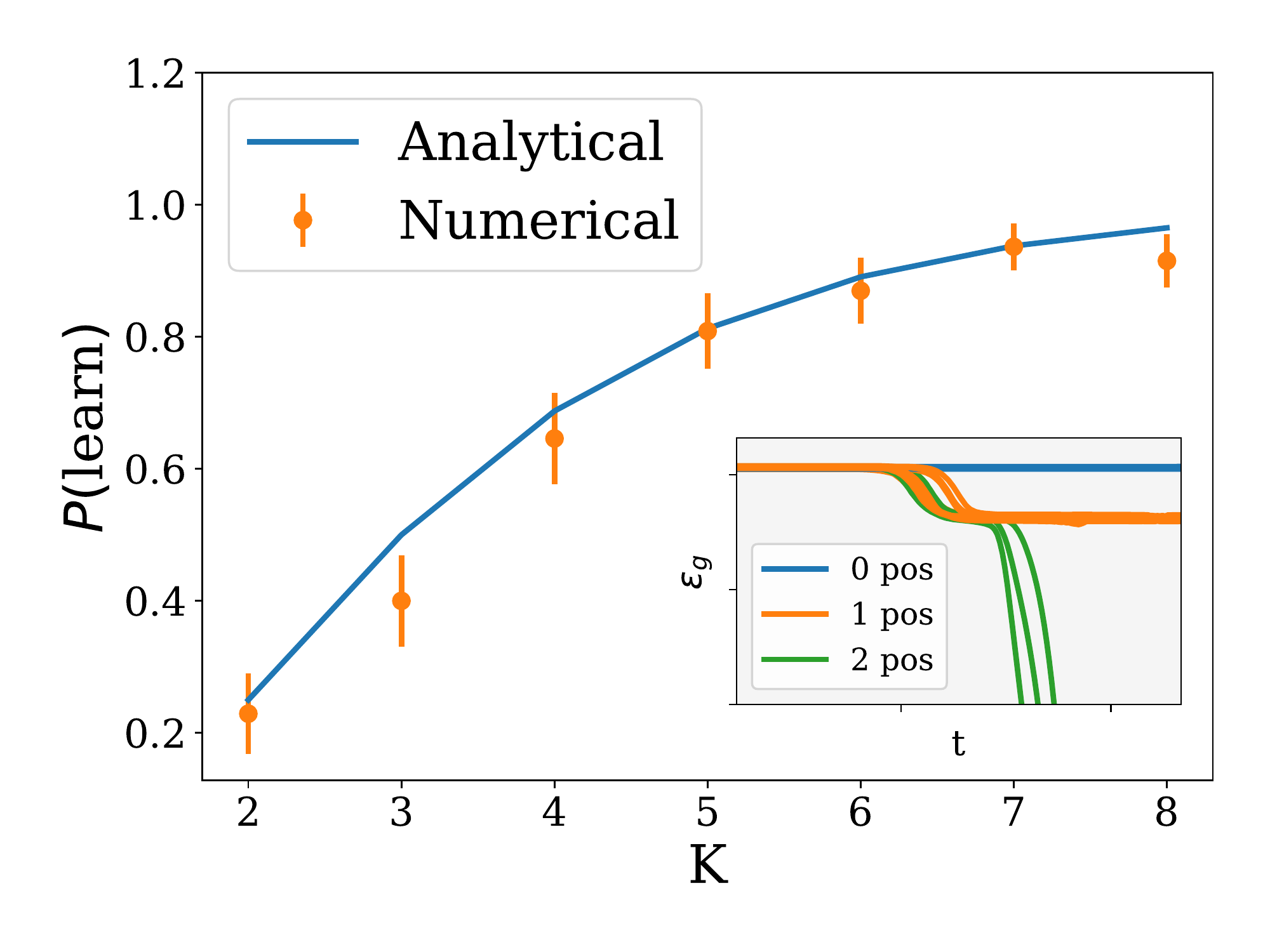}%
  \caption{ \label{fig:relu} \textbf{ Over-parameterisation improves performance
      of shallow ReLU networks.} We show the learning dynamics of a student with
    $K=3$ hidden nodes trained on a teacher with $M=2$ nodes and
    $\tilde W_2^m=1$ if the feedback vector has \textcolor{C0}{0},
    \textcolor{C1}{1}, or \textcolor{C2}{2} positive entries. \textit{Inset}:
    Probability of achieving zero test error (Eq.~\ref{eq:p_learn}, line)
    compared to the fraction of simulations that converged to zero test error
    (out of 50, crosses). \emph{Other parameters}:
    $N=500, \eta=0.1, \sigma_0=10^{-2}$.}
\end{figure}

The ReLU activation function possesses the continuous symmetry
$\max(0, x)=\gamma \max(0, x / \gamma)$ for any $\gamma>0$ preventing ReLU networks to compensate a change of sign of $W_2^k$ with a change of sign of
$W_1^k$. Consequently, a ReLU student can only simultaneously align to the feedback
vector $F_1$ and recover the teacher's second layer $\tilde W_2$ if at least $M$
elements of $F_1$ have the same sign as $\tilde W_2$.  The inset of Fig.~\ref{fig:relu} shows that a student trained on a teacher with $M=2$ second-layer
weights $\tilde W_2^m=1$ only converges to zero test error if the feedback
vector has 2 positive elements (green). If instead the feedback vector has only
0 (blue) or 1 (orange) positive entry, the student will settle at a finite test
error.  More generally, the probability of perfect recovery
for a student with $K\geq M$ nodes sampled randomly is given analytically as:
\begin{equation}
  \label{eq:p_learn}
  P(\text{learn})=\frac{1}{2^K} \sum_{k=0}^M \binom{K}{k}.
\end{equation}
As shown in Fig.~\ref{fig:relu}, this formula matches with simulations. Note that the importance of the ``correct'' sign for the feedback
matrices was also observed in deep neural networks
by~\citet{liao2016important}.

\subsection{Degeneracy breaking in deep networks}
\label{sec:db-deep}

\begin{figure*}[t!]
  \centering
  \includegraphics[width=.55\linewidth]{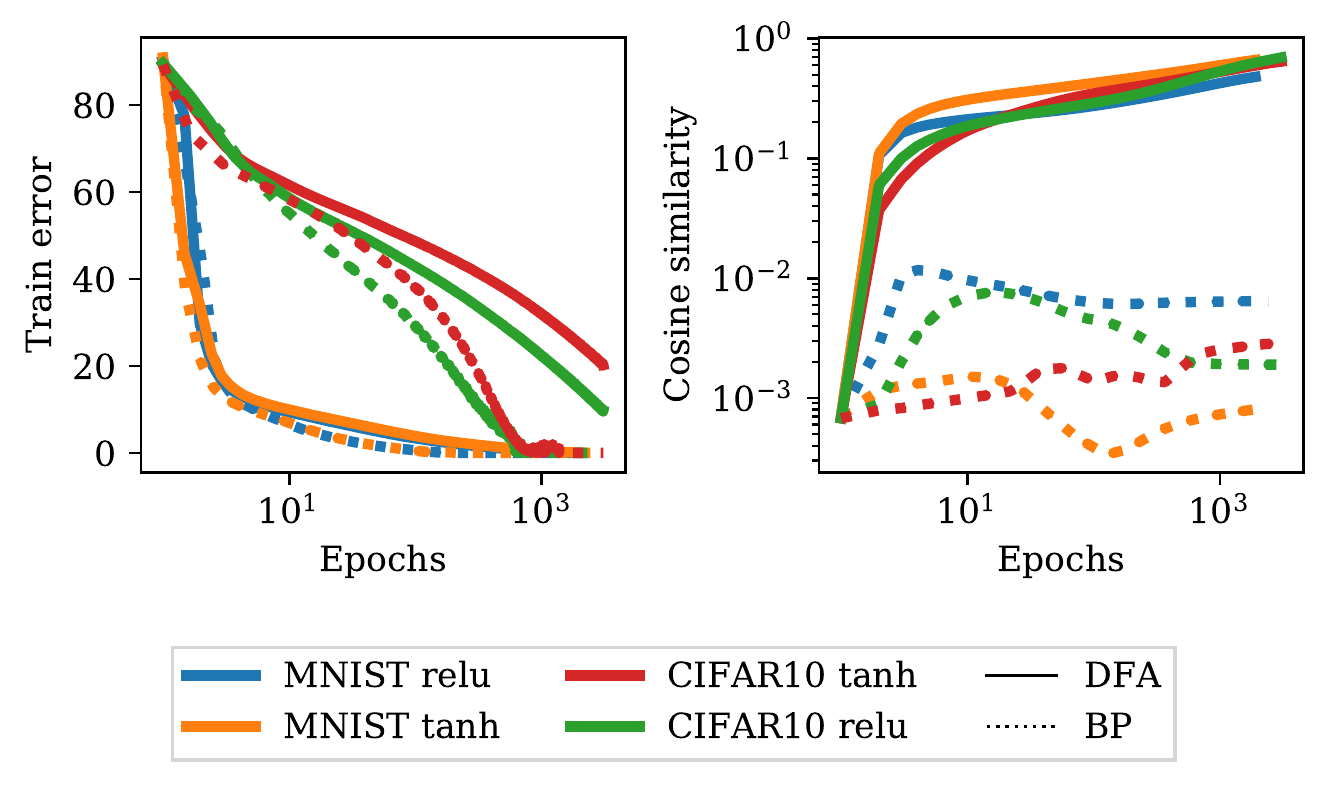}\qquad%
  \includegraphics[width=.4\linewidth]{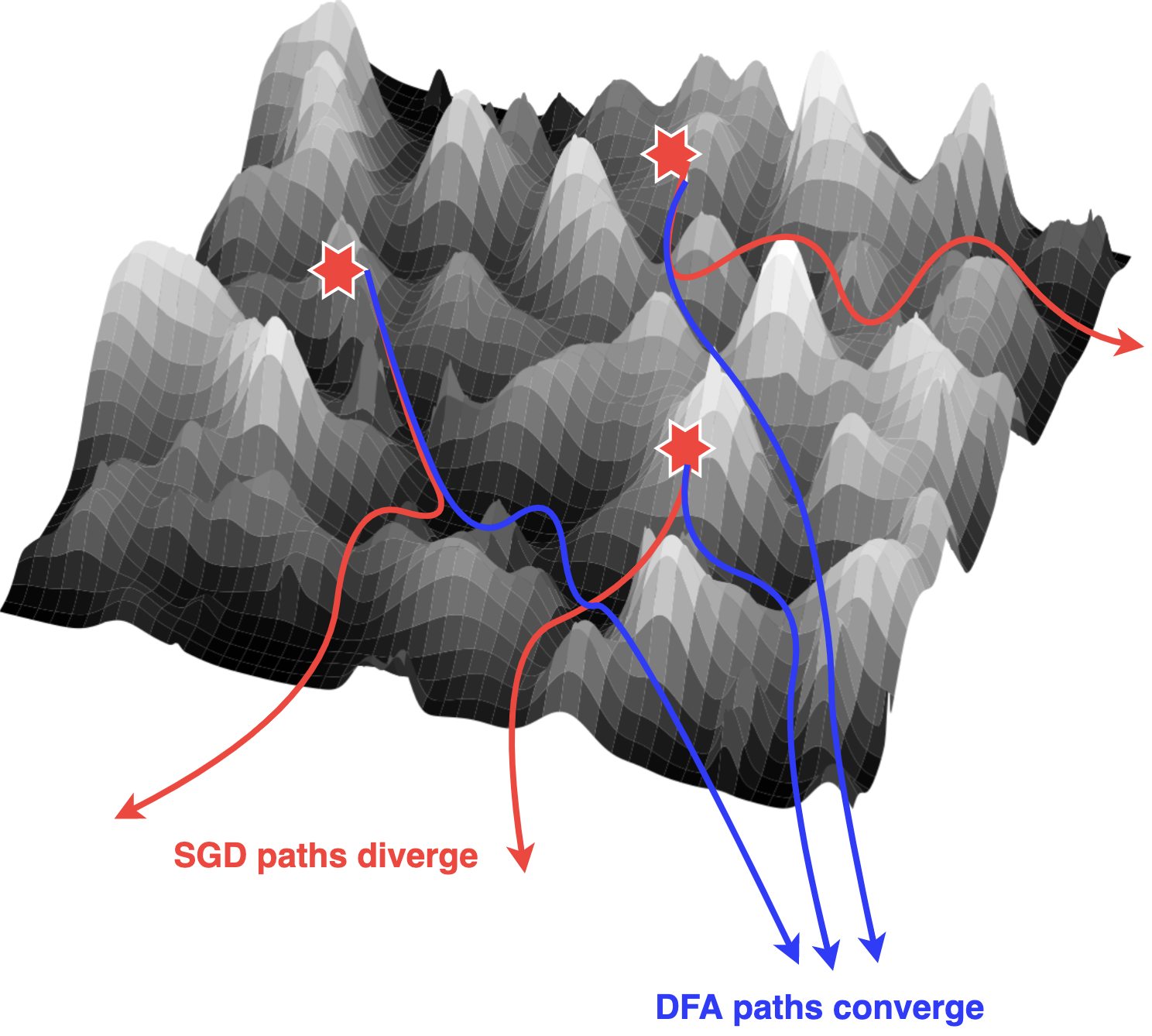}
  \caption{\label{fig:db-deep}\textbf{Degeneracy breaking also occurs in deep
      neural networks.} (\emph{Left}) We plot the training accuracy and the
    cosine similarity between the weights of four-layer fully-connected neural
    networks with sigmoidal and ReLU activations during training on MNIST and
    CIFAR10. Averages taken over 10 runs; for exp.~details see
    Sec.~\ref{sec:db-deep}. (\emph{Right}) Cartoon of the degeneracy breaking
    process in the loss landscape of a deep network: while the optimization paths of models trained with SGD diverge in the loss landscape, with DFA they converge to a region of the landscape determined by the feedback matrices.}
\end{figure*}

We explore to what extent degeneracy breaking occurs in deep nonlinear networks by training 4-layer multi-layer perceptrons (MLPs) with
100 nodes per layer for 1000 epochs with both BP and DFA, on the MNIST and
CIFAR10 datasets, with Tanh and ReLU nonlinearities
(cf. App.~\ref{app:degeneracy} for further experimental details). The dynamics of the training
loss, shown in the left of Fig.~\ref{fig:db-deep}, are very similar for BP and DFA.

From degeneracy breaking, one expects DFA to drive the optimization path towards a special region of the loss landscape determined by the feedback matrices. We test this hypothesis by measuring whether networks trained with the same feedback matrices from different initial weights converge towards the same region of the landscape.
The cosine similarity between the vectors obtained by stacking the weights of two networks trained independently
using BP reaches at most $10^{-2}$ (right of Fig.~\ref{fig:db-deep}), signalling that they reach very distinct minima. In contrast, when trained with DFA, networks reach a cosine similarity between $0.5$ and $1$ at convergence, thereby confirming that DFA breaks the degeneracy between the solutions in the landscape and biases towards a special region of the loss landscape, both for sigmoidal and ReLU activation functions.

This result suggests that heavily over-parametrised neural networks used in practice can be trained
successfully with DFA because they have a large number of degenerate
solutions. We leave a more detailed exploration of the interplay between DFA and the loss landscape for future work.
As we discuss in Sec.~\ref{sec:nonlinear} the Align-then-Memorise mechanism sketched in Fig.~\ref{fig:degeneracy_breaking}~c also occurs in deep non-linear
networks. 

\section{How do gradients align in deep networks?}
\label{sec:wa}

This section focuses on the alignment phase of learning. In the
two-layer setup there is a single feedback vector $F_1$,
of same dimensions as the second layer $W_2$, and to which $W_2$ must align
in order for the first layer to recover the true gradient. 

In deep networks, as each layer $W_l$ has a distinct feedback matrix $F_l$ of different size of $W_l$, it is not obvious
how the weights must align to ensure gradient alignment. 
We study how the alignment occurs by considering
deep linear networks with $L$ layers without bias, without any assumption
on the training data. 
While the expressivity of linear networks is naturally
limited, their learning dynamics is non-linear and rich enough to give insights
that carry over to the non-linear case both for BP~\cite{baldi1989neural,
  le1991eigenvalues, krogh1992generalization, saxe2014exact, advani2020high} and
for DFA~\cite{lillicrap2016random, noekland2016direct, frenkel2019learning}.

\subsection{Weight alignment as a natural structure}

In the following, we assume that the weights are initialised to zero.
With BP, they would stay zero at all times, but for DFA the layers become nonzero sequentially, from the bottom to the top
layer.
In the linear setup, the updates of the first two layers at time $t$ can be written in
terms of the corresponding input and error vectors using Eq.~\eqref{eq:dfa-update}\footnote{We implicitly assume
  a minibatch size of 1 for notational simplicity, but conclusions are unchanged
  in the finite-batch setup.}:
\begin{align}
  \delta W_{1}^t=-\eta (F_1 e_t) x_t^{T},\quad  \delta W_{2}^t=-\eta (F_2 e_t) {(W_1 x_t)}^\top
\label{eq:linear-updates}
\end{align}
Summing these updates shows that the first layer performs Hebbian learning
modulated by the feedback matrix $F_1$:
\begin{align}
  \label{eq:A1}
  W_1^t &= -\eta \sum\nolimits_{t'=0}^{t-1} F_1 e_{t'} x_{t'}^\top = F_1 A_1^t, \\
  A_1^t &= -\eta \sum\nolimits_{t'=0}^{t-1} e_{t'} x_{t'}^\top
\end{align}
Plugging this result into $\delta W_2^t$, we obtain:
\begin{align}
    W_2^t &= -\eta \sum_{t'=0}^{t-1} F_2 e_t (A_1^{t'} x_{t'})^\top F_1^\top = F_2 A_2^t F_1^\top, \\
    A_2^t &= \eta^2 \sum_{t'=0}^{t-1}\sum_{t''=0}^{t'-1} (x_{t'}\cdot x_{t''}) e_{t'}e_{t''}^\top. 
\end{align}
When iterated, the procedure above reveals that DFA naturally leads to \emph{weak weight alignment} of the
network weights to the feedback matrices:
\begin{align}
  \label{eq:weak-wa}
  \text{\textbf{Weak WA: }} W_{1<l<L}^{t}\! =\! F_l A_l^t F_{l-1}^\top\!, \quad W_L^{t}\! = \!A_L^t F_{L-1}^\top,
\end{align}
where we defined the \emph{alignment matrices}
$A_{l\geq 2}^t\in \mathbb{R}^{n_L\times n_L}$:
\begin{align}
\label{eq:a-formula}
    A_{l\geq 2}^t = \eta^2 \sum_{t'=0}^{t-1} \sum_{t''=0}^{t'-1} (B_{l}^{t'} x_{t'})\cdot(B_{l}^{t''} x_{t''}) e_{t'} e_{t''}^\top.
\end{align}
$B_{l}\in \mathbb{R}^{n_L\times n_L}$ is defined recursively as a function
of the feedback matrices only and its expression together with the full derivation is deferred to App.~\ref{app:proof}.
These results can be adapted both to DRTP~\cite{frenkel2019learning}, another variant of feedback
alignment where $e_t=-y_t$ and to FA by
performing the replacement $F_l\to F_l F_{l+1} \ldots F_{L-1}$.

\subsection{Weight alignment leads to gradient alignment}
Weak WA builds throughout training, but does not directly imply GA. 
However, if the alignment matrices become proportional to the identity, we obtain \emph{strong weight alignment}:
\begin{equation}
  \label{eq:strong-wa}
  \mathrm{\textbf{Strong WA:}}\quad W_{1<l<L}^{t} \propto F_l
  F_{l-1}^\top, \quad
  W_{L}^{t} \propto F_{L-1}^\top.
\end{equation}

Additionally, since GA requires $F_l e \propto W_{l+1}^\top \delta a_{l+1}$ (Eqs.~\ref{eq:dfa-update} and~\ref{eq:bp-update}), strong WA directly implies GA if the feedback matrices $F_{l \ge 2}$ are assumed left-orthogonal, i.e. $F_l^\top F_l = \mathbb I_{n_L}$. 
Strong WA of~\eqref{eq:strong-wa} induces the weights, by the orthogonality condition, to cancel out by pairs of two:
\begin{align}
    W_{l+1}^\top \delta a_{l+1} &\propto F_{l} F_{l+1}^\top F_{l+1} \ldots F_{L-1}^\top F_{L-1} e = F_{l} e.
\end{align}
The above suggests that taking the feedback matrices left-orthogonal is favourable for GA. If the feedback matrices elements are sampled i.i.d.\ from a
Gaussian distribution, GA still holds in expectation since $\mathbb{E} \left[F_{l}^\top F_{l}\right] \propto \mathbb I_{n_L}$.

\paragraph{Quantifying gradient alignment}
Our analysis shows that key to GA are the alignment matrices: the closer they are to identity, i.e.~the better their conditioning, the
stronger the GA. This comes at the price of restricted
expressivity, since layers are encouraged to align to a product of (random)
feedback matrices. 
In the extreme case of strong WA,  the freedom of layers $l\geq2$ is entirely sacrificed to allow learning in the first layer! 
This is not harmful for the linear networks as the first layer alone is
enough to maintain full expressivity\footnote{such an alignment was indeed
  already observed in the linear setup for BP~\cite{ji2018gradient}.}.
Nonlinear networks, as argued in Sec.~\ref{sec:db}, rely on the Degeneracy Breaking mechanism
to recover expressivity.

\section{The case of deep nonlinear networks}
\label{sec:nonlinear}

\begin{figure}[t!]
    \centering
    \includegraphics[width=\columnwidth]{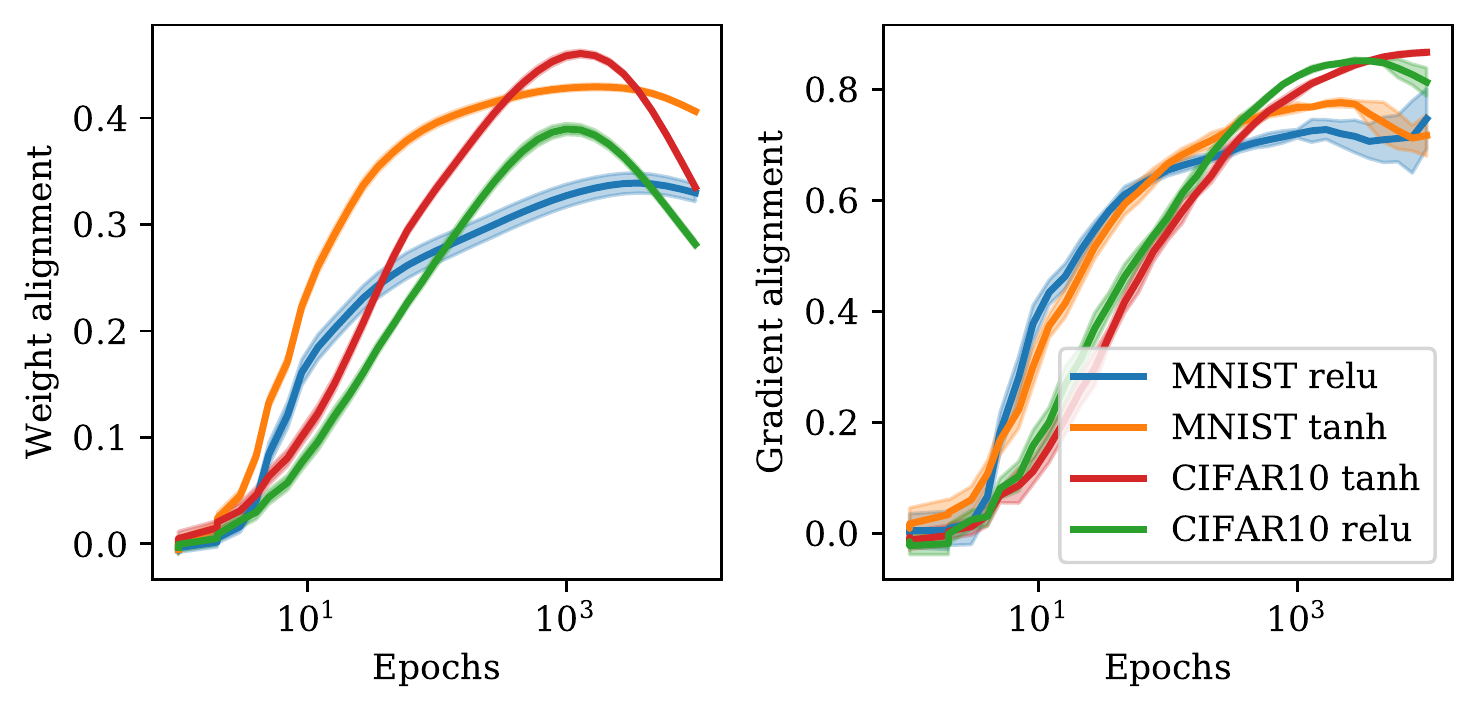}    
    \caption{\textbf{Global alignment dynamics of deep nonlinear networks exhibits Align-then-Memorise.} Global weight and gradient alignments, as defined in~\eqref{eq:wa-global}, varying the activation function and the dataset. Shaded regions represent the (small) variability over 10 runs.}
    \label{fig:wa-global}
\end{figure}

\begin{figure}[t!]
  \centering
  \includegraphics[width=\columnwidth]{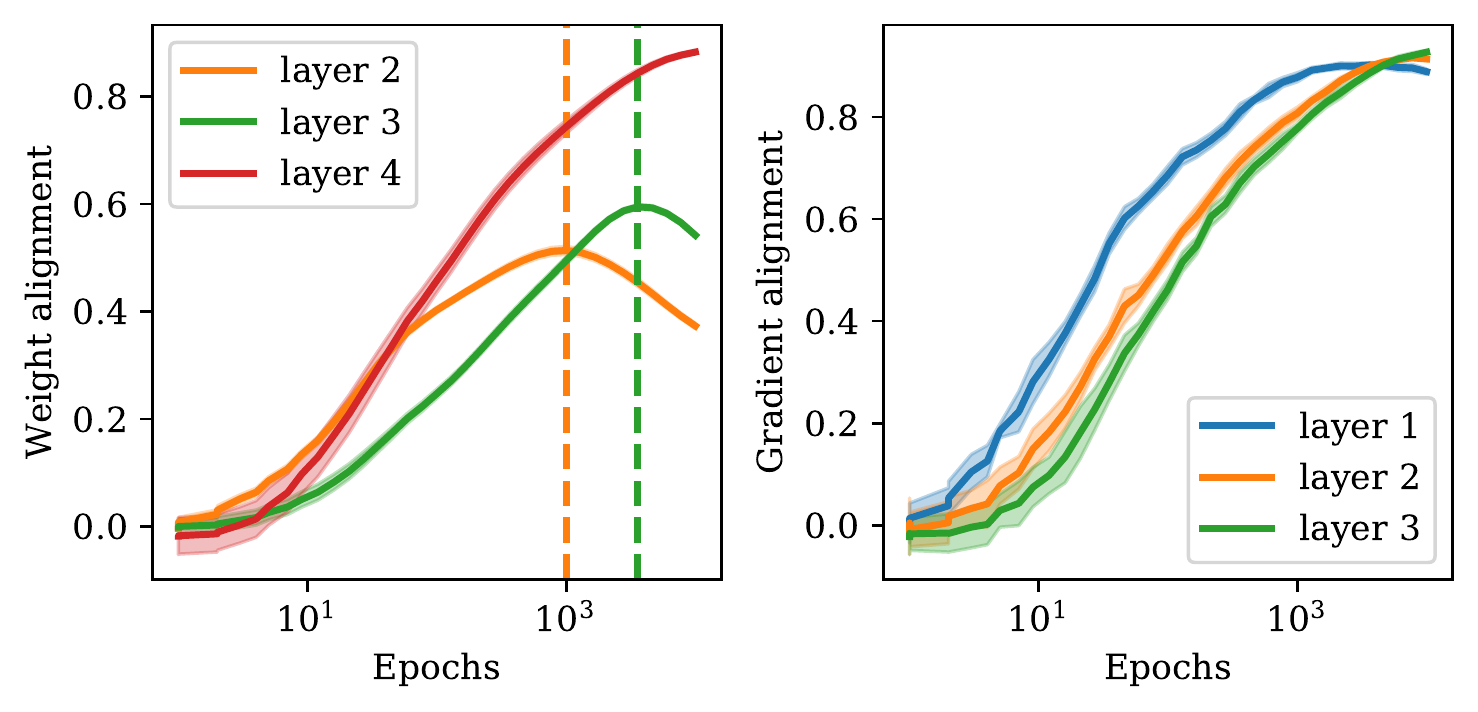}
    \caption{\textbf{Layerwise alignment dynamics reveal sequential Align-then-Memorise.} Layerwise weight and gradient alignments as defined in~\eqref{eq:wa-layerwise}, for a ReLU network trained on CIFAR10 with 10\% label corruption. Shaded regions represent the (small) variability over 10 runs.}
    \label{fig:wa-layerwise}
\end{figure}

In this section, we show that the theoretical predictions of the previous two sections hold remarkably well in deep nonlinear
networks trained on standard vision datasets.

\subsection{Weight Alignment occurs like in the linear setup}

To determine whether WA described in Sec.~\ref{sec:wa} holds in
the deep nonlinear setup of Sec.~\ref{sec:db-deep}, we introduce the global and layerwise alignment
observables:
\begin{align}
    &\mathrm{WA}\! =\! \measuredangle \left(\mathbf{F}, \mathbf{W} \right)\!,
    &&\mathrm{GA} \!= \!\measuredangle \left( \mathbf{G}^\mathrm{DFA}, \mathbf{G}^\mathrm{BP} \right)
    \label{eq:wa-global}\\
    &\mathrm{WA}_{l\geq 2}\! =\! \measuredangle\! \left(\mathbf{F}_l\!,\! \mathbf{W}_l \right)\!,
    &&\mathrm{GA}_{l\geq 2}\! = \!\measuredangle \left( \mathbf{G}^\mathrm{DFA}_l\!,\! \mathbf{G}^\mathrm{BP}_l \right),
    \label{eq:wa-layerwise}
\end{align}
where $\measuredangle(\mathbf{A}, \mathbf{B}) = \mathrm{Vec}(\mathbf{A})\cdot \mathrm{Vec}(\mathbf{B})/ \Vert \mathbf{A} \Vert \Vert \mathbf{B}\Vert$ and
\begin{align*}
    \mathbf{F} &= \left(F_2 F_1^\top, \ldots, F_{L-1} F_{L-2}^\top,  F_{L-1}^\top\right),\\
    \mathbf{W}(t) &= \left(W_2^t, \ldots, W_{L-1}^t, W_L^t\right),\\
    \mathbf{G}(t) &= \left(\delta a_1^t, \ldots, \delta a_{L-1}^t\right).
\end{align*}
Note that the layer-wise alignment of $W_l$ with $F_l F_{l-1}^\top$ was never measured before: it differs from the alignment of $F_l$ with $W_{l+1} \ldots W_L$ observed in~\cite{crafton2019direct}, which is more akin to GA. 

If $\mathbf{W}$ and $\mathbf{F}$ were uncorrelated, the WA defined in~\eqref{eq:wa-global} would be vanishing as the width of the layer grows large. Remarkably, WA becomes of order one after a few epochs as shown in Fig.~\ref{fig:wa-global} (left), and
strongly correlates with GA (right). This suggests that the
layer-wise WA uncovered for linear networks with weights initialized to zero also drives GA in the general case.

\subsection{Align-then-Memorise occurs from bottom layers to top}

As can be seen in Fig.~\ref{fig:wa-global}, WA clearly reaches a maximum then decreases, as expected from the Align-then-Memorise process. Notice that the decrease is stronger for CIFAR10 than it is for MNIST, since CIFAR-10 is much harder to fit than MNIST: more WA needs to be sacrificed. Increasing label corruption similarly makes the datasets harder to fit,
and decreases the final WA, as detailed in SM~\ref{app:degeneracy}. However,
another question arises: why does the GA keep increasing in this case, in spite
of the decreasing WA?

To answer this question, we need to disentangle the dynamics of the layers of
the network, as in Eq.~\eqref{eq:wa-layerwise}. In Fig.~\ref{fig:wa-layerwise}, we focus on the ReLU network
applied to CIFAR10, and shuffle 10\% of the labels in the training set to make the
Align-then-Memorise procedure more easily visible. Although the network contains 4
layers of weights, we only have 3 curves for WA and GA: WA is only defined for
layers 2 to 4 according to Eq.~\eqref{eq:wa-layerwise}, whereas GA of the last layer is not represented here since it is always equal to one.

As can be seen, the second layer is the first to start aligning: it reaches its
maximal WA around $1000$ epochs (orange dashed line), then decreases. The third layer starts aligning later
and reaches its maximal WA around $2000$ epochs (green dashed line), then decreases. As for the last
layer, the WA is monotonically increasing. Hence, the Align-then-Memorise
mechanism operates in a layerwise fashion, starting from the bottom layers to
the top layers.

Note that the WA of the last layers is the most crucial, since it affects the GA
of all the layers below, whereas the WA of the second layer only affects the GA
of the first layer. It therefore makes sense to keep the WA of the last layers
high, and let the bottom layers perform the memorization first. This is
reminiscent of the linear setup, where all the layers align except for the
first, which does all the learning. In fact, this strategy enables the GA of
each individual layer to keep increasing until late times: the diminishing WA of
the bottom layers is compensated by the increasing WA of the top layers.





\section{What can hamper alignment?}
\label{sec:when}

\begin{figure}[tb]
\centering
  \includegraphics[width=.6\columnwidth]{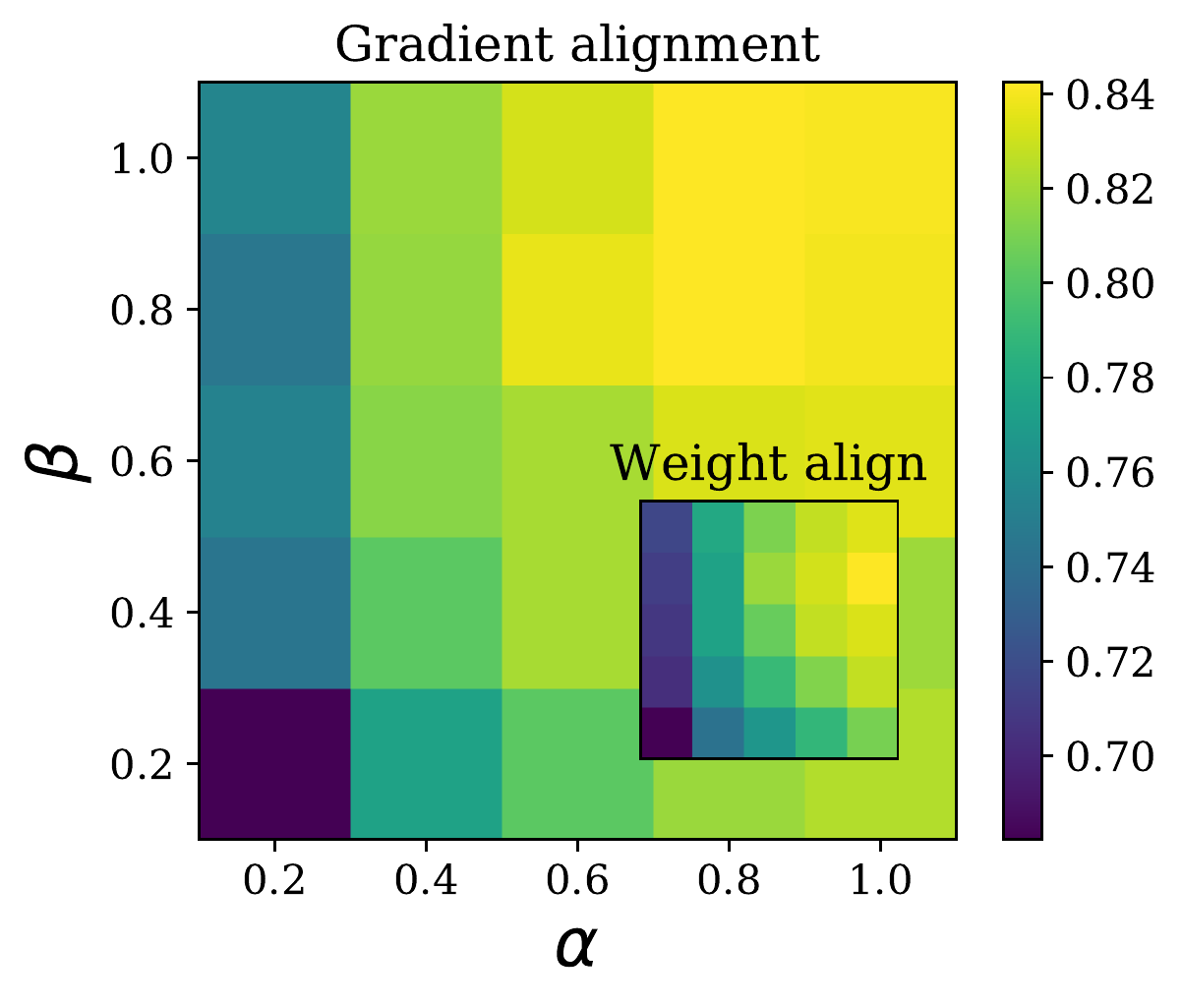}
  \caption{ \label{fig:alphabeta} \textbf{Badly conditioned output statistics
      can hamper alignment.} WA and GA at the final point of training decrease
    when the output classes are correlated ($\beta<1$) or of different variances
    ($\alpha<1$).}
\end{figure}

\begin{figure}[tb]
  \centering
  \includegraphics[width=\columnwidth]{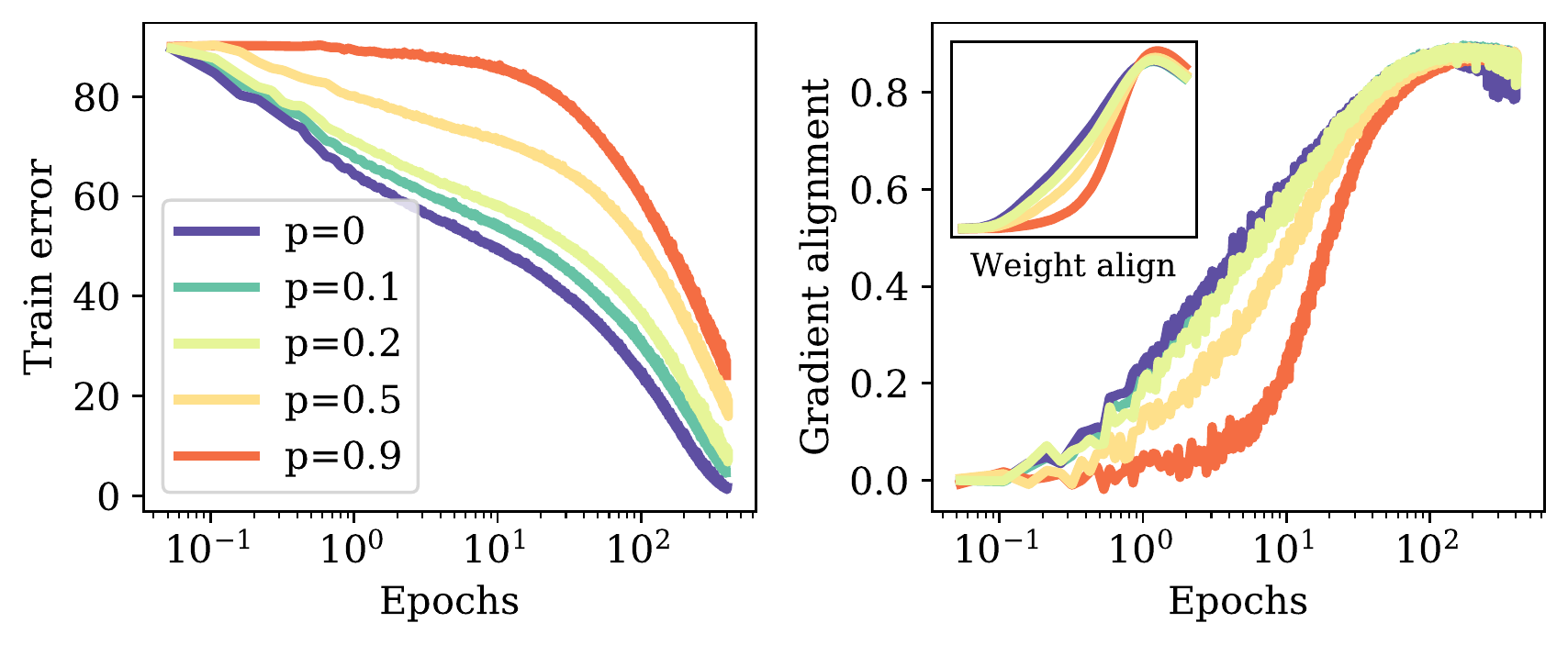}
  \caption{\textbf{Label corruption hampers alignment in the early stages of training.} We see that the higher the label corruption, the more time WA and GA take to start increasing, since the network initially predicts equal probabilities over the output classes.}
  \label{fig:noise}
\end{figure}

We demonstrated that GA is enabled by the WA mechanism, both theoretically for
linear networks and numerically for nonlinear networks. In this section, we
leverage our analysis of WA to identify situations in which GA fails.

\subsection{Alignment is data-dependent}
\label{sec:data-dependence}

In the linear case, GA occurs if the alignment matrices presented in
Sec.~\ref{sec:wa} are well conditioned. Note that if the output size $n_L$ is
equal to one, e.g.\ for scalar regression or binary classification tasks, then
the alignment matrices are simply scalars, and GA is guaranteed. When this is
not the case, one can obtain the deviation from GA by studying the expression of
the alignment matrices~\eqref{eq:a-formula}. They are formed by summing outer
products of the error vectors $e_{t'} e_{t''}^\top$, where
$e_t = \hat y_t\!-\!y_t$. Therefore, good conditioning requires the different
components of the errors to be uncorrelated and of similar variances. This can
be violated by (i) the targets $y$, or (ii) the predictions $\hat y$.

\paragraph{(i) Structure of data} The first scenario can be demonstrated
in a simple regression task on i.i.d.\ Gaussian inputs $x\sim \mathbb R^{10}$. The
targets $y\in \mathbb R^2$ are randomly sampled from the following distribution:
\begin{align}
    y\!\sim\! \mathcal{N}(0,\Sigma), \quad
    \Sigma\! =\! \begin{pmatrix}
    1 & \alpha(1-\beta)\\
    \alpha(1-\beta) & \alpha^2
    \end{pmatrix}\!,\!\! \quad \alpha, \beta\!\leq 1.
\end{align}
In Fig.~\ref{fig:alphabeta}, we show the final WA and GA of a 3-layer ReLU
network trained for $10^3$ epochs on $10^3$ examples sampled from this
distribution (further details in SM~\ref{app:targets}). As predicted, imbalanced
($\alpha<1$) or correlated ($\beta<1$) target statistics hamper WA and GA. Note
that the inputs also come into play in Eq.~\eqref{eq:a-formula}: a more
detailed theoretical analysis of the impact of input and target statistics on
alignment is deferred to SM~\ref{app:drtp}.

\paragraph{(ii) Effect of noise}
For classification tasks, the targets $y$ are one-hot encodings whose statistics
are naturally well conditioned. However, alignment can be
degraded if the statistics of the predictions $\hat y$ become correlated.

One can enforce such a correlation in CIFAR10 by shuffling a fraction $p$ of the
labels. The WA and GA dynamics of a 3-layer ReLU network are shown in
Fig.~\ref{fig:noise}. At high $p$, the
network can only perform random guessing during the first few epochs, and
assigns equal probabilities to the 10 classes. The correlated structure of the predictions prevents
alignment until the network starts to fit the random labels: the predictions of
the different classes then decouple and WA takes off, leading to GA. 

\subsection{Alignment is impossible for convolutional layers}

A convolutional layer with filters $H_l$ can be represented by a large
fully-connected layer whose weights are represented by a block Toeplitz matrix
$\phi(H_l)$~\cite{d2019finding}. This matrix has repeated blocks due to weight
sharing, and most of its weights are equal to zero due to locality. In order to
verify WA and therefore GA, the following condition must hold:
$\phi(H_l)\propto F_l F_{l-1}^\top$. Yet, due to the very constrained structure
of $\phi(H_l)$, this is impossible for a general choice of $F_l$. Therefore, the WA mechanism suggests
a simple explanation for why GA doesn't occur in vanilla CNNs, and confirms
the previously stated hypothesis that CNNs don't have enough flexibility to
align~\cite{launay2019principled}.

In the case of convolutional layers, this lack of alignment makes learning near
to impossible, and has lead practitioners to design
alternatives~\cite{han2019direct, moskovitz2018feedback}. However, the extent to
which alignment correlates with good performance in the general setup (both in
terms of fitting and generalisation) is a complex question which we leave for
future work. Indeed, nothing prevents DFA from finding a good optimization path,
different from the one followed by BP. Conversely, obtaining high gradient
alignment at the end of training is not a sufficient condition for DFA to
retrieve the results of BP, e.g.\ if the initial trajectory leads to a wrong
direction.


\section*{Acknowledgements}

We thank Florent Krzakala for introducing us to feedback alignment, and we thank
him and Lenka Zdeborov\'a for organising the Les Houches 2020 workshop on
Statistical Physics and Machine Learning where this work was initiated. We thank
Florent Krzakala, Giulio Biroli, Charlotte Frenkel, Julien Launay, Martin
Lefebvre, Leonardo Petrini, Iacopo Poli, Levent Sagun and Mihiel Straat for
helpful discussions. MR acknowledges funding from the French Agence Nationale de la Recherche under grant ANR-19-P3IA-0001 PRAIRIE. SD acknowledges funding from PRAIRIE for a visit to Trieste
to collaborate on this project. RO acknowledges funding from the Region
Ile-de-France.

\bibliography{refs}

\begin{thebibliography}{48}
\providecommand{\natexlab}[1]{#1}
\providecommand{\url}[1]{\texttt{#1}}
\expandafter\ifx\csname urlstyle\endcsname\relax
  \providecommand{\doi}[1]{doi: #1}\else
  \providecommand{\doi}{doi: \begingroup \urlstyle{rm}\Url}\fi

\bibitem[Advani et~al.(2020)Advani, Saxe, and Sompolinsky]{advani2020high}
Advani, M.~S., Saxe, A.~M., and Sompolinsky, H.
\newblock High-dimensional dynamics of generalization error in neural networks.
\newblock \emph{Neural Networks}, 132:\penalty0 428 -- 446, 2020.

\bibitem[Aubin et~al.(2018)Aubin, Maillard, Barbier, Krzakala, Macris, and
  Zdeborov{\'{a}}]{aubin2018committee}
Aubin, B., Maillard, A., Barbier, J., Krzakala, F., Macris, N., and
  Zdeborov{\'{a}}, L.
\newblock {The committee machine: Computational to statistical gaps in learning
  a two-layers neural network}.
\newblock In \emph{Advances in Neural Information Processing Systems 31}, pp.\
  3227--3238, 2018.

\bibitem[Bahri et~al.(2020)Bahri, Kadmon, Pennington, Schoenholz,
  Sohl-Dickstein, and Ganguli]{bahri2020statistical}
Bahri, Y., Kadmon, J., Pennington, J., Schoenholz, S., Sohl-Dickstein, J., and
  Ganguli, S.
\newblock {Statistical Mechanics of Deep Learning}.
\newblock \emph{Annual Review of Condensed Matter Physics}, 11\penalty0
  (1):\penalty0 501--528, 2020.

\bibitem[Baity-Jesi et~al.(2018)Baity-Jesi, Sagun, Geiger, Spigler, Arous,
  Cammarota, LeCun, Wyart, and Biroli]{baity-jesi2018}
Baity-Jesi, M., Sagun, L., Geiger, M., Spigler, S., Arous, G., Cammarota, C.,
  LeCun, Y., Wyart, M., and Biroli, G.
\newblock {Comparing Dynamics: Deep Neural Networks versus Glassy Systems}.
\newblock In \emph{Proceedings of the 35th International Conference on Machine
  Learning}, 2018.

\bibitem[Baldi \& Hornik(1989)Baldi and Hornik]{baldi1989neural}
Baldi, P. and Hornik, K.
\newblock Neural networks and principal component analysis: Learning from
  examples without local minima.
\newblock \emph{Neural networks}, 2\penalty0 (1):\penalty0 53--58, 1989.

\bibitem[Bartunov et~al.(2018)Bartunov, Santoro, Richards, Marris, Hinton, and
  Lillicrap]{bartunov2018assessing}
Bartunov, S., Santoro, A., Richards, B., Marris, L., Hinton, G.~E., and
  Lillicrap, T.
\newblock Assessing the scalability of biologically-motivated deep learning
  algorithms and architectures.
\newblock In \emph{Advances in Neural Information Processing Systems}, pp.\
  9368--9378, 2018.

\bibitem[Biehl \& Schwarze(1995)Biehl and Schwarze]{biehl1995}
Biehl, M. and Schwarze, H.
\newblock {Learning by on-line gradient descent}.
\newblock \emph{J. Phys. A. Math. Gen.}, 28\penalty0 (3):\penalty0 643--656,
  1995.

\bibitem[Brutzkus \& Globerson(2017)Brutzkus and
  Globerson]{brutzkus2017globally}
Brutzkus, A. and Globerson, A.
\newblock Globally optimal gradient descent for a convnet with gaussian inputs.
\newblock In \emph{Proceedings of the 34th International Conference on Machine
  Learning - Volume 70}, ICML'17, pp.\  605–614, 2017.

\bibitem[Chizat \& Bach(2018)Chizat and Bach]{chizat2018}
Chizat, L. and Bach, F.
\newblock On the global convergence of gradient descent for over-parameterized
  models using optimal transport.
\newblock In \emph{Advances in Neural Information Processing Systems 31}, pp.\
  3040--3050, 2018.

\bibitem[Crafton et~al.(2019)Crafton, Parihar, Gebhardt, and
  Raychowdhury]{crafton2019direct}
Crafton, B., Parihar, A., Gebhardt, E., and Raychowdhury, A.
\newblock Direct feedback alignment with sparse connections for local learning.
\newblock \emph{Frontiers in neuroscience}, 13:\penalty0 525, 2019.

\bibitem[Crick(1989)]{crick1989recent}
Crick, F.
\newblock The recent excitement about neural networks.
\newblock \emph{Nature}, 337\penalty0 (6203):\penalty0 129--132, 1989.

\bibitem[d'Ascoli et~al.(2019)d'Ascoli, Sagun, Biroli, and Bruna]{d2019finding}
d'Ascoli, S., Sagun, L., Biroli, G., and Bruna, J.
\newblock Finding the needle in the haystack with convolutions: on the benefits
  of architectural bias.
\newblock In \emph{Advances in Neural Information Processing Systems}, pp.\
  9334--9345, 2019.

\bibitem[Du et~al.(2018)Du, Lee, Tian, Singh, and Poczos]{du2018gradient}
Du, S., Lee, J., Tian, Y., Singh, A., and Poczos, B.
\newblock Gradient descent learns one-hidden-layer {CNN}: Don’t be afraid of
  spurious local minima.
\newblock In \emph{Proceedings of the 35th International Conference on Machine
  Learning}, volume~80, pp.\  1339--1348, 2018.

\bibitem[Engel \& Van~den Broeck(2001)Engel and Van~den
  Broeck]{engel2001statistical}
Engel, A. and Van~den Broeck, C.
\newblock \emph{Statistical mechanics of learning}.
\newblock Cambridge University Press, 2001.

\bibitem[Frenkel et~al.(2019)Frenkel, Lefebvre, and Bol]{frenkel2019learning}
Frenkel, C., Lefebvre, M., and Bol, D.
\newblock Learning without feedback: Direct random target projection as a
  feedback-alignment algorithm with layerwise feedforward training.
\newblock 2019.

\bibitem[Gabri{\'{e}}(2020)]{gabrie2020meanfield}
Gabri{\'{e}}, M.
\newblock Mean-field inference methods for neural networks.
\newblock \emph{Journal of Physics A: Mathematical and Theoretical},
  53\penalty0 (22):\penalty0 223002, 2020.

\bibitem[Gardner \& Derrida(1989)Gardner and Derrida]{gardner1989}
Gardner, E. and Derrida, B.
\newblock {Three unfinished works on the optimal storage capacity of networks}.
\newblock \emph{Journal of Physics A: Mathematical and General}, 22\penalty0
  (12):\penalty0 1983--1994, 1989.

\bibitem[Ghorbani et~al.(2019)Ghorbani, Mei, Misiakiewicz, and
  Montanari]{ghorbani2019limitations}
Ghorbani, B., Mei, S., Misiakiewicz, T., and Montanari, A.
\newblock Limitations of lazy training of two-layers neural network.
\newblock In \emph{Advances in Neural Information Processing Systems 32}, pp.\
  9111--9121, 2019.

\bibitem[Gilmer et~al.(2017)Gilmer, Raffel, Schoenholz, Raghu, and
  Sohl-Dickstein]{gilmer2017explaining}
Gilmer, J., Raffel, C., Schoenholz, S.~S., Raghu, M., and Sohl-Dickstein, J.
\newblock Explaining the learning dynamics of direct feedback alignment.
\newblock In \emph{ICLR workshop track}, 2017.

\bibitem[Goldt et~al.(2019)Goldt, Advani, Saxe, Krzakala, and
  Zdeborov{\'a}]{goldt2019dynamics}
Goldt, S., Advani, M., Saxe, A., Krzakala, F., and Zdeborov{\'a}, L.
\newblock Dynamics of stochastic gradient descent for two-layer neural networks
  in the teacher-student setup.
\newblock In \emph{Advances in Neural Information Processing Systems 32}, 2019.

\bibitem[Grossberg(1987)]{grossberg1987competitive}
Grossberg, S.
\newblock Competitive learning: From interactive activation to adaptive
  resonance.
\newblock \emph{Cognitive science}, 11\penalty0 (1):\penalty0 23--63, 1987.

\bibitem[Han \& Yoo(2019)Han and Yoo]{han2019direct}
Han, D. and Yoo, H.-j.
\newblock Direct feedback alignment based convolutional neural network training
  for low-power online learning processor.
\newblock In \emph{Proceedings of the IEEE International Conference on Computer
  Vision Workshops}, 2019.

\bibitem[Ji \& Telgarsky(2019)Ji and Telgarsky]{ji2018gradient}
Ji, Z. and Telgarsky, M.
\newblock Gradient descent aligns the layers of deep linear networks.
\newblock In \emph{International Conference on Learning Representations
  (ICLR)}, 2019.

\bibitem[Kinzel \& Ruj{\'{a}}n(1990)Kinzel and Ruj{\'{a}}n]{kinzel1990}
Kinzel, W. and Ruj{\'{a}}n, P.
\newblock {Improving a Network Generalization Ability by Selecting Examples}.
\newblock \emph{EPL (Europhysics Letters)}, 13\penalty0 (5):\penalty0 473--477,
  1990.

\bibitem[Krogh \& Hertz(1992)Krogh and Hertz]{krogh1992generalization}
Krogh, A. and Hertz, J.~A.
\newblock Generalization in a linear perceptron in the presence of noise.
\newblock \emph{Journal of Physics A: Mathematical and General}, 25\penalty0
  (5):\penalty0 1135, 1992.

\bibitem[Launay et~al.(2019)Launay, Poli, and Krzakala]{launay2019principled}
Launay, J., Poli, I., and Krzakala, F.
\newblock Principled training of neural networks with direct feedback
  alignment.
\newblock \emph{arXiv:1906.04554}, 2019.

\bibitem[Launay et~al.(2020)Launay, Poli, Boniface, and
  Krzakala]{launay2020direct}
Launay, J., Poli, I., Boniface, F., and Krzakala, F.
\newblock Direct feedback alignment scales to modern deep learning tasks and
  architectures.
\newblock In \emph{Advances in neural information processing systems}, 2020.

\bibitem[Le~Cun et~al.(1991)Le~Cun, Kanter, and Solla]{le1991eigenvalues}
Le~Cun, Y., Kanter, I., and Solla, S.~A.
\newblock Eigenvalues of covariance matrices: Application to neural-network
  learning.
\newblock \emph{Physical Review Letters}, 66\penalty0 (18):\penalty0 2396,
  1991.

\bibitem[Liao et~al.(2016)Liao, Leibo, and Poggio]{liao2016important}
Liao, Q., Leibo, J.~Z., and Poggio, T.
\newblock How important is weight symmetry in backpropagation?
\newblock In \emph{Proceedings of the Thirtieth AAAI Conference on Artificial
  Intelligence}, pp.\  1837--1844, 2016.

\bibitem[Lillicrap et~al.(2016)Lillicrap, Cownden, Tweed, and
  Akerman]{lillicrap2016random}
Lillicrap, T., Cownden, D., Tweed, D., and Akerman, C.
\newblock {Random synaptic feedback weights support error backpropagation for
  deep learning}.
\newblock \emph{Nature Communications}, 7:\penalty0 1--10, 2016.

\bibitem[Mei et~al.(2018)Mei, Montanari, and Nguyen]{mei2018}
Mei, S., Montanari, A., and Nguyen, P.
\newblock {A mean field view of the landscape of two-layer neural networks}.
\newblock \emph{Proceedings of the National Academy of Sciences}, 115\penalty0
  (33):\penalty0 E7665--E7671, 2018.

\bibitem[Moskovitz et~al.(2018)Moskovitz, Litwin-Kumar, and
  Abbott]{moskovitz2018feedback}
Moskovitz, T.~H., Litwin-Kumar, A., and Abbott, L.
\newblock Feedback alignment in deep convolutional networks.
\newblock \emph{arXiv preprint arXiv:1812.06488}, 2018.

\bibitem[N{\o}kland(2016)]{noekland2016direct}
N{\o}kland, A.
\newblock {Direct Feedback Alignment Provides Learning in Deep Neural
  Networks}.
\newblock In \emph{Advances in Neural Information Processing Systems 29}, 2016.

\bibitem[Rotskoff \& Vanden-Eijnden(2018)Rotskoff and
  Vanden-Eijnden]{rotskoff2018}
Rotskoff, G. and Vanden-Eijnden, E.
\newblock {Parameters as interacting particles: long time convergence and
  asymptotic error scaling of neural networks}.
\newblock In \emph{Advances in Neural Information Processing Systems 31}, pp.\
  7146--7155, 2018.

\bibitem[Rumelhart et~al.(1986)Rumelhart, Hinton, and
  Williams]{rumelhart1986learning}
Rumelhart, D.~E., Hinton, G.~E., and Williams, R.~J.
\newblock Learning representations by back-propagating errors.
\newblock \emph{Nature}, 323\penalty0 (6088):\penalty0 533--536, 1986.

\bibitem[Saad(2009)]{saad2009line}
Saad, D.
\newblock \emph{On-line learning in neural networks}, volume~17.
\newblock Cambridge University Press, 2009.

\bibitem[Saad \& Solla(1995{\natexlab{a}})Saad and Solla]{saad1995a}
Saad, D. and Solla, S.
\newblock {Exact Solution for On-Line Learning in Multilayer Neural Networks}.
\newblock \emph{Phys. Rev. Lett.}, 74\penalty0 (21):\penalty0 4337--4340,
  1995{\natexlab{a}}.

\bibitem[Saad \& Solla(1995{\natexlab{b}})Saad and Solla]{saad1995b}
Saad, D. and Solla, S.
\newblock {On-line learning in soft committee machines}.
\newblock \emph{Phys. Rev. E}, 52\penalty0 (4):\penalty0 4225--4243,
  1995{\natexlab{b}}.

\bibitem[Saxe et~al.(2014)Saxe, McClelland, and Ganguli]{saxe2014exact}
Saxe, A., McClelland, J., and Ganguli, S.
\newblock {Exact solutions to the nonlinear dynamics of learning in deep linear
  neural networks}.
\newblock In \emph{International Conference on Learning Representations
  (ICLR)}, 2014.

\bibitem[Saxe et~al.(2018)Saxe, Bansal, Dapello, Advani, Kolchinsky, Tracey,
  and Cox]{saxe2018information}
Saxe, A., Bansal, Y., Dapello, J., Advani, M., Kolchinsky, A., Tracey, B., and
  Cox, D.
\newblock {On the information bottleneck theory of deep learning}.
\newblock In \emph{ICLR}, 2018.

\bibitem[Seung et~al.(1992)Seung, Sompolinsky, and Tishby]{seung1992}
Seung, H.~S., Sompolinsky, H., and Tishby, N.
\newblock {Statistical mechanics of learning from examples}.
\newblock \emph{Physical Review A}, 45\penalty0 (8):\penalty0 6056--6091, 1992.

\bibitem[Sirignano \& Spiliopoulos(2019)Sirignano and
  Spiliopoulos]{sirignano2018}
Sirignano, J. and Spiliopoulos, K.
\newblock {Mean field analysis of neural networks: A central limit theorem}.
\newblock \emph{Stochastic Processes and their Applications}, 2019.

\bibitem[Soltanolkotabi et~al.(2018)Soltanolkotabi, Javanmard, and
  Lee]{soltanolkotabi2018theoretical}
Soltanolkotabi, M., Javanmard, A., and Lee, J.
\newblock Theoretical insights into the optimization landscape of
  over-parameterized shallow neural networks.
\newblock \emph{IEEE Transactions on Information Theory}, 65\penalty0
  (2):\penalty0 742--769, 2018.

\bibitem[Tian(2017)]{tian2017analytical}
Tian, Y.
\newblock An analytical formula of population gradient for two-layered relu
  network and its applications in convergence and critical point analysis.
\newblock In \emph{Proceedings of the 34th International Conference on Machine
  Learning (ICML)}, pp.\  3404–3413, 2017.

\bibitem[Watkin et~al.(1993)Watkin, Rau, and Biehl]{watkin1993}
Watkin, T., Rau, A., and Biehl, M.
\newblock {The statistical mechanics of learning a rule}.
\newblock \emph{Reviews of Modern Physics}, 65\penalty0 (2):\penalty0 499--556,
  1993.

\bibitem[Yoshida \& Okada(2019)Yoshida and Okada]{yoshida2019datadependence}
Yoshida, Y. and Okada, M.
\newblock Data-dependence of plateau phenomenon in learning with neural network
  --- statistical mechanical analysis.
\newblock In \emph{Advances in Neural Information Processing Systems 32}, pp.\
  1720--1728, 2019.

\bibitem[Zdeborov{\'{a}} \& Krzakala(2016)Zdeborov{\'{a}} and
  Krzakala]{zdeborova2016}
Zdeborov{\'{a}}, L. and Krzakala, F.
\newblock {Statistical physics of inference: thresholds and algorithms}.
\newblock \emph{Adv. Phys.}, 65\penalty0 (5):\penalty0 453--552, 2016.

\bibitem[Zhong et~al.(2017)Zhong, Song, Jain, Bartlett, and
  Dhillon]{zhong2017recovery}
Zhong, K., Song, Z., Jain, P., Bartlett, P., and Dhillon, I.
\newblock Recovery guarantees for one-hidden-layer neural networks.
\newblock In \emph{Proceedings of the 34th International Conference on Machine
  Learning-Volume 70}, pp.\  4140--4149, 2017.

\end{thebibliography}
\bibliographystyle{icml2020}

\clearpage
\appendix
\section{Derivation of the ODE}
\label{sec:EOM}
The derivation of the ODE's that describe the dynamics of the test error for
shallow networks closely follows the one of~\citet{saad1995a} and
\citet{biehl1995} for back-propagation. Here, we give the main steps to obtain
the analytical curves of the main text and refer the reader to their paper for
further details.

As we discuss in Sec.~\ref{sec:db}, student and teacher are both two-layer
networks with $K$ and $M$ hidden nodes, respectively. For an input
$x\in\mathbb R^N$, their outputs $\hat y$ and $y$ can be written as
\begin{align}
    \hat y &= \phi_\theta (x) = \sum_{k=1}^K W_2^k\, g \left( \lambda^k\right), \notag\\ y&=\phi_{\tilde \theta} (x)=\sum_{m=1}^M \tilde W_2^m\, g \left( \nu^m\right),
\end{align}
where we have introduced the pre-activations
$\lambda^k \equiv W_1^k x / \sqrt{N}$ and
$\nu^m \equiv \tilde W_1^m x / \sqrt{N}$. Evaluating the test error of a student
with respect to the teacher under the squared loss leads us to compute the
average
\begin{equation}
  \label{eq:supp_eg-teacher-student}
  \epsilon_g \left(\theta, \tilde{\theta}\right) = \frac{1}{2} \EE_x \left[
    \sum_{k=1}^K W_2^k\, g \left( \lambda^k\right)- \sum_{m=1}^M \tilde W_2^m\, g \left( \nu^m\right)\right]^2,
\end{equation}
where the expectation is taken over inputs $x$ for a fixed student and
teacher. Since $x$ only enters Eq.~\eqref{eq:supp_eg-teacher-student} via the
pre-activations $\lambda = (\lambda^k)$ and $\nu = (\nu^m)$, we can replace the
high-dimensional average over $x$ by a low-dimensional average over the $K+M$
variables $(\lambda, \nu)$. The pre-activations are jointly Gaussian since the
inputs are drawn element-wise i.i.d.\ from the Gaussian distribution. The mean
of~$(\lambda, \nu)$ is zero since $\EE x_i=0$, so the distribution of
$(\lambda, \nu)$ is fully described by the second moments
\begin{gather}
  Q^{kl}= \EE \lambda^k \lambda^l = W_1^k\cdot W_1^l/N,\\
  R^{km}= \EE \lambda^k \nu^m = W_1^k\cdot\tilde W_1^m/N, \\
  T^{mn}=\EE \nu^m \nu^n = \tilde W_1^m\cdot\tilde W_1^n / N.
\end{gather}
which are the ``order parameters'' that we introduced in the main text. We can
thus rewrite the generalisation error~\eqref{eq:eg} as a function of only the
order parameters and the second-layer weights,
\begin{equation}
  \label{eq:supp_eg}
  \lim_{N\to\infty} \epsilon_g(\theta, \tilde \theta)
  = \epsilon_g(Q, R, T, W_2, \tilde{W}_2)
\end{equation}
As we update the weights using SGD, the time-dependent order parameters $Q, R$,
and $W_2$ evolve in time. By choosing different scalings for the learning rates
in the SGD updates~\eqref{eq:dfa-update}, namely
$$\eta_{W_1}=\eta,\qquad \eta_{W_2}=\eta/N$$ for some constant $\eta$, we
guarantee that the dynamics of the order parameters can be described by a set of
ordinary differential equations, called their ``equations of motion''.  We can
obtain these equations in a heuristic manner by squaring the weight
update~\eqref{eq:dfa-update} and taking inner products with $\tilde{W}_1^m$, to
yield the equations of motion for $Q$ and $R$ respectively:
\begin{subequations}\label{eq:eom}
  \begin{align}
    \frac{\dd R^{km}}{\dd \alpha} &= -\eta F_1^k \EE \left[g'(\lambda^k) \nu^m
                                    e\right] \\
    \frac{\dd Q^{k\ell}}{\dd \alpha} &= - \eta F_1^k \EE \left[g'(\lambda^k)\lambda^\ell e\right] - \eta F_1^\ell \EE\left[ g'(\lambda^\ell) \lambda^k e\right]\notag\\
    &+ \eta^2 F_1^k F_1^\ell \EE\left[g'(\lambda^k) g'(\lambda^\ell) e^2\right],\\
    \frac{\dd W_2^{k}}{\dd \alpha} &= - \eta \EE \left[g(\lambda^k) e\right]
  \end{align}
\end{subequations}
where, as in the main text, we introduced the error
$e=\phi_\theta(x)-\phi_{\tilde \theta}(x)$.  In the limit $N\to\infty$, the
variable $\alpha=\mu / N$ becomes a continuous time-like variable. The remaining
averages over the pre-activations, such as
$$\EE g'(\lambda^k)\lambda^\ell g(\nu^m),$$ are simple three-dimensional
integral over the Gaussian random variables $\lambda^k, \lambda^\ell$ and
$\nu^m$ and can be evaluated analytically for the choice of
$g(x)=\erf(x/\sqrt{2})$~\cite{biehl1995} and for linear networks with
$g(x)=x$. Furthermore, these averages can be expressed only in term of the order
parameters, and so the equations close. We note that the asymptotic exactness of
Eqs.~\ref{eq:eom} can be proven using the techniques used recently to prove the
equations of motion for BP~\cite{goldt2019dynamics}.

We provide an integrator for the full system of ODEs for any $K$ and $M$ in the
Github
repository. 

\section{Detailed analysis of DFA dynamics}

In this section, we present a detailed analysis of the ODE dynamics in the
matched case $K=M$ for sigmoidal networks
($g(x)=\erf\left( \nicefrac{x}{\sqrt{2}} \right)$).

\paragraph{The Early Stages and Gradient Alignment}
\label{sec:early-stages}
We now use Eqs.~\eqref{eq:eom} to demonstrate that alignment occurs in the early
stages of learning, determining from the start the solution DFA will converge to
(see Fig.~\ref{fig:degeneracy_breaking} which summarises the dynamical evolution
of the student's second layer weights).

Assuming zero initial weights for the student and orthogonal first layer weights
for the teacher (i.e.~$T^{nm}$ is the identity matrix), for small times
($t\ll 1$), one can expand the order parameters in $t$:
\begin{align}
    R^{km}(t)&=t\dot{R}^{km}(0)+\mathcal{O}(t^2),\notag\\ Q^{kl}(t)&=t\dot{Q}^{kl}(0)+\mathcal{O}(t^2),\notag\\
    W_2^{k}(t)&=t\dot{W}_2^{k}(0)+\mathcal{O}(t^2).
    \label{eq:B11}
\end{align}
where, due to the initial conditions, $R(0)=Q(0)=W_2(0)=0$. 
Using  Eq.~\ref{eq:eom}, we can obtain the lowest order term of the above updates:
\begin{align}
\label{eq:derivs}
    \dot{R}^{km}(0)&=\frac{\sqrt{2}}{\pi}\eta \tilde{W}_2^m F_1^k, \notag\\
  \dot{Q}^{kl}(0)&=\frac{2}{\pi}\eta^2 \left((\tilde{W}_2^{k})^2+(\tilde{W}_2^{l})^2\right) F_1^l F_1^k,\notag\\
    \dot{W}_2^{k}(0)&=0
\end{align}
Since both $\dot{R}(0)$ and $\dot Q(0)$ are non-zero, this initial condition is not a fixed point of DFA.
To analyse initial alignment, we consider the first order term of $\dot{W}_2$. Using Eq.~\eqref{eq:B11} with the derivatives at $t=0$~\eqref{eq:derivs}, we obtain to linear order in $t$:
\begin{align}
\dot{W}_2^{k}(t)
=& \frac{2}{\pi^2}\eta^2 ||\tilde{W}_2||^2 F_1^k t.
\end{align}
Crucially, this update is in the direction of the feedback vector $F_1$. DFA training thus constrains the student to initially grow in the direction of the feedback vector and align with it. This implies gradient alignment between BP and DFA and dictates into which of the many degenerate solutions in the energy landscape the student converges.

\paragraph{Plateau phase}
After the initial phase of learning with DFA where the test error decreases
exponentially, similarly to BP, the student falls into a symmetric fixed point
of the Eqs.~(\ref{eq:eom}) where the weights of a single student node are
correlated to the weights of all the teacher nodes (\cite{saad1995a, biehl1995,
  engel2001statistical}). The test error stays constant while the student is
trapped in this fixed point.  We can obtain an analytic expression for the order
parameters under the assumption that the teacher first-layer weights are
orthogonal ($T^{nm}=\delta_{nm}$). We set the teacher's second-layer weights to
unity for notational simplicity ($\tilde{W}_2^m=1$) and restrict to linear order
in the learning rate $\eta$, since this is the dominant contribution to the
learning dynamics at early times and on the plateau~\cite{saad1995b}. In the
case where all components of the feedback vector are positive, the order
parameters are of the form $Q^{kl}=q, R^{km}=r, W_2^k=w_2$ with:
\begin{align}
    q=\frac{1}{2K-1},\quad
    r=\sqrt{\frac{q}{2}},\quad
    w_2=\sqrt{\frac{1+2q}{q(4+3q)}}.
\end{align}
If the components of the feedback vector are not all positive, we instead obtain $R^{km} =\operatorname{sgn}(F^k) r$, $W_2^{k} =\operatorname{sgn}(F^k) w_2$ and $Q^{kl} =\operatorname{sgn}(F^k)\operatorname{sgn}(F^l) q$. This shows that on the plateau the student is already in the configuration that maximises its alignment with $F_1$.
Note that in all cases, the value of the test error reached at the plateau is the same for DFA and BP.

\paragraph{Memorisation phase and Asymptotic Fixed Point}
At the end of the plateau phase, the student converges to its final solution, which is often referred to as the \emph{specialised} phase~\cite{saad1995a, biehl1995, engel2001statistical}.
The configuration of the order parameters is such that the student reproduces her teacher up to sign changes that guarantee the alignment between $W_2$ and $F_1$ is maximal, i.e.~$\operatorname{sgn}(W^k_2)=\operatorname{sgn}(F_1^k)$. 
The final value of the test error of a student trained with DFA is the same as that of a student trained with BP on the same teacher.

\paragraph{Choice of the feedback vector}

\begin{figure}[h!]
    \centering
    \includegraphics[width=.8\linewidth]{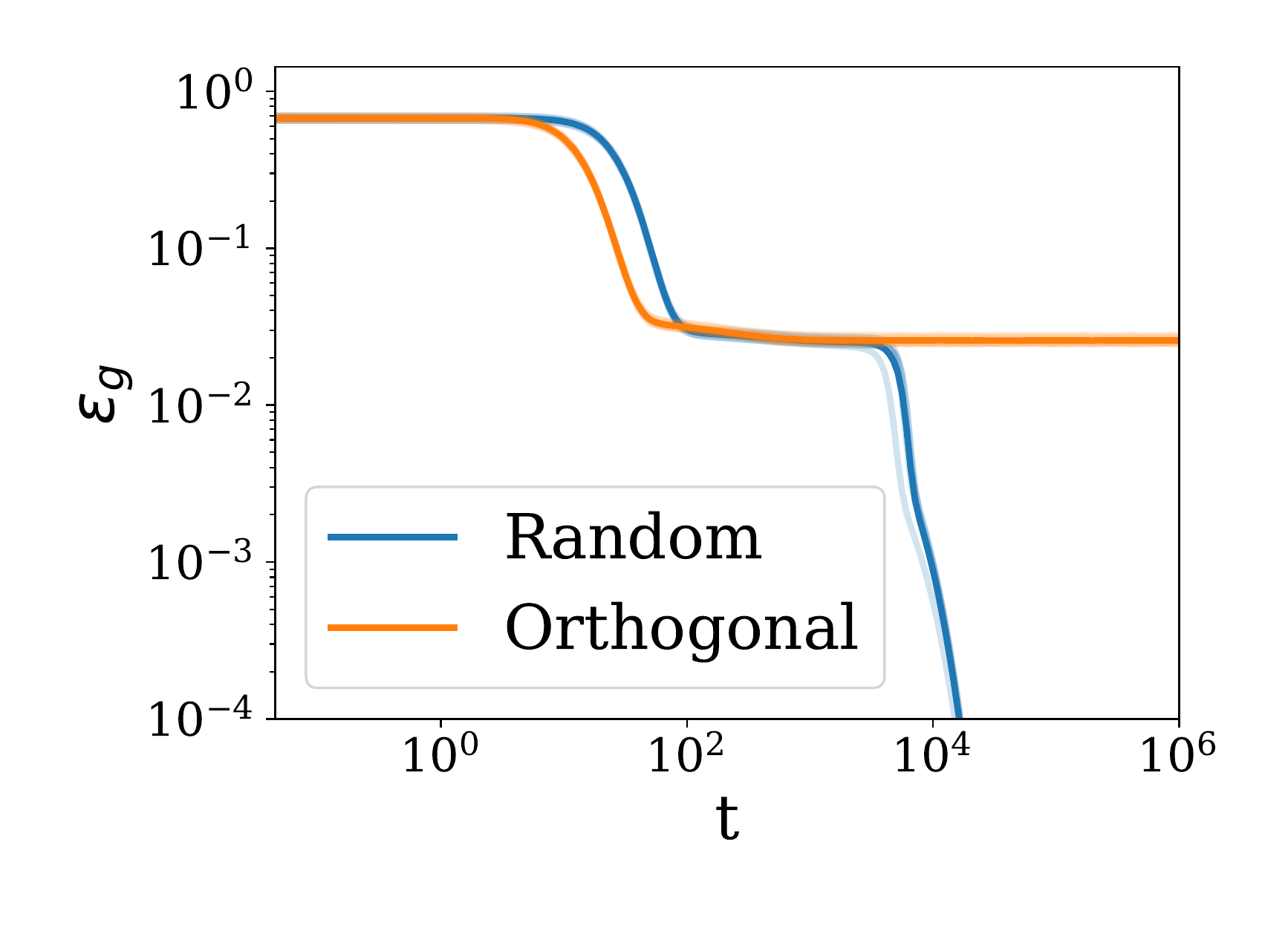}
    \caption{Test error of a sigmoidal student started with zero initial weights. The feedback vector $F_1$ is chosen random (\textcolor{C0}{blue}) and orthogonal to the teacher's second layer weights $\tilde W_2$ (\textcolor{C1}{orange}). \emph{Parameters:}
      $\eta=0.1, K=M=2$.  }
    \label{fig:evil_B}
\end{figure}
In the main text, we saw how a wrong choice of feedback vector $F_1$ can prevent
a ReLU student from learning a task. Here, we show that also for sigmoidal
student, a \emph{wrong} choice of feedback vector
$F_1$ is possible. As Fig.~\ref{fig:evil_B} shows, in the case where the $F_1$
is taken orthogonal to the teacher second layer weights, a student whose weights
are initialised to zero remains stuck on the plateau and is unable to
learn. In contrast, when the $F_1$ is chosen with random i.i.d. components drawn from the standard normal distribution, perfect recovery is achieved.

\section{Derivation of weight alignment}
\label{app:proof}

Since the network is linear, the update equations are (consider the first three
layers only):
\begin{align}
\delta W_{1}&=-\eta (F_1 e) x^{T}, \\
\quad \delta W_{2}&=-\eta (F_2 e) (W_1 x)^\top, \\
\delta W_{3}&=-\eta (F_3 e) (W_2 W_1 x)^\top
\end{align}
First, it is straightforward to see that 
\begin{align}
    W_1^t &= -\eta \sum_{t'=0}^{t-1} F_1 e_{t'} x_{t'}^\top = F_1 A_1^t\\
    A_1^t &= -\eta \sum_{t'=0}^{t-1} e_{t'} x_{t'}^\top
\end{align}
This allows to calculate the dynamics of $W_2^t$:
\begin{align}
    \delta W_2^t &= -\eta F_2 e_t (A_1^{t} x_t)^\top F_1^\top \\
    W_2^t &= -\eta \sum_{t'=0}^{t-1} F_2 e_t (A_1^{t'} x_{t'})^\top F_1^\top = F_2 A_2^t F_1^\top\\
    A_2^t &= -\eta \sum_{t'=0}^{t-1} e_{t'} (A_1^{t'} x_{t'})^\top = \eta^2 \sum_{t'=0}^{t-1}\sum_{t''=0}^{t'-1} (x_{t'}\cdot x_{t''}) e_{t'}e_{t''}^\top. 
\end{align}
Which in turns allows to calculate the dynamics of $W_3^t$:
\begin{align}
    \delta W_3^t &= -\eta F_3 e_t (F_2 A_2^{t'} F_1^\top F_1 A_1^{t'} x_t)^\top \\
    W_3^t &= -\eta \sum_{t'=0}^{t-1} F_3 e_{t'} (F_2 A_2^{t'} F_1^\top F_1 A_1^{t'} x_t)^\top = F_3 A_3^t F_2^\top\\
    A_3^t &= -\eta \sum_{t'=0}^{t-1} F_3 e_{t'} (A_2^{t'} F_1^\top F_1 A_1^{t'} x_{t'})^\top \\
    &= \eta^2 \sum_{t'=0}^{t-1}\sum_{t''=0}^{t'-1} (A_1^{t'} x_{t'}) \cdot (A_1^{t''} x_{t''}) e_{t'}e_{t''}^\top. 
\end{align} 
By induction it is easy to show the general expression:
\begin{align}
    A_1^t &= -\eta \sum_{t'=0}^{t-1} e_{t'} x_{t'}^\top \\
    A_2^t &= \eta^2 \sum_{t'=0}^{t-1}\sum_{t''=0}^{t'-1} (x_{t'}\cdot x_{t''}) e_{t'}e_{t''}^\top\\
    A_{l\geq 3}^t &= \eta^2 \sum_{t,t'=0} (A_{l-2}^{t'} \ldots A_{1}^{t'} x_{t'})\cdot(A_{l-2}^{t''} \ldots A_{1}^{t''} x_{t''}) e_{t'} e_{t''}^\top
\end{align}

Defining $A_0 \equiv \mathbb{I}_{n_0}$, one can rewrite this as in Eq.~\ref{eq:a-formula}
\begin{align}
    A_{l\geq 2}^t &= \eta^2 \sum_{t'=0}^{t-1} \sum_{t''=0}^{t'-1} (B_l^{t'} x_{t'})\cdot(B_l^{t''} x_{t''}) e_{t'} e_{t''}^\top, \\
    B_l &= A_{l-2} \cdots A_{0}.
\end{align}


\section{Impact of data structure}
\label{app:drtp}

To study the impact of data structure on the alignment, the simplest setup to
consider is that of Direct Random Target
Projection~\cite{frenkel2019learning}. Indeed, in this case the error vector
$e_t = -y_t$ does not depend on the prediction of the network: the dynamics
become explicitly solvable in the linear case.

For concreteness, we consider the setup of~\cite{lillicrap2016random} where the
targets are given by a linear teacher, $y = Tx$, and the inputs are i.i.d
Gaussian. We denote the input and target correlation matrices as follows:
\begin{align}
    \mathbb E\left[x x^\top\right] \equiv \Sigma_{x} \in\mathbb R^{n_0\times n_0}, \\
    \mathbb E\left[T T^\top\right] \equiv \Sigma_{y}\in\mathbb R^{n_L\times n_L}
\end{align}
If the batch size is large enough, one can write $x_t x_t^\top = \mathbb E\left[x x^\top\right] = \Sigma_{x}$. Hence the dynamics of Eq.~\ref{eq:linear-updates} become:
\begin{align}
  \delta W_{1}^t&=-\eta (F_1 e_t) x_t^{T} = \eta F_1 T x_t x_t^\top = \eta F_1 T \Sigma_{x}\\
  \delta W_{2}^t&= - \eta (F_2 e_t) {(W_1 x_t)}^\top = \eta F_2 T \Sigma_x W_1^\top \\&=  \eta^2 F_2 \left(T \Sigma_{x}^2 T^\top\right) F_1^\top\\
  \delta W_{3}^t&= - \eta (F_3 e_t) {(W_2 W_1 x_t)}^\top = \eta F_3 T \Sigma_x W_1^\top W_2^\top \\&=  \eta^3 F_3 \left(T \Sigma_{x}^2 T^\top\right) \left(T \Sigma_x^2 T^\top\right) F_2^\top
\end{align}

From which we easily deduce $A_1^t = \eta T \Sigma_{x} t$, and the expression of the alignment matrices at all times:
\begin{align}
    A_{l\geq 2}^t &= \eta^l\left( T \Sigma_{x}^2 T^\top\right)^{l-1} t
\end{align}

As we saw, GA depends on how well-conditioned the alignement matrices are, i.e. how different it is from the identity. To examine deviation from identity, we write $\Sigma_x = \mathbb I_{n_0} + \tilde \Sigma_x$ and $\Sigma_y = \mathbb I_{n_L} + \tilde \Sigma_y$, where the tilde matrices are small perturbations. Then to first order,
\begin{align}
    A_{l\geq 2}^t - I_{n_L} &\propto \mathbb  (l-1) \left( \tilde\Sigma_y + 2 T\tilde\Sigma_x T^\top \right)
\end{align}

Here we see that GA depends on how well-conditioned the input and target correlation matrices $\Sigma_x$ and $\Sigma_y$ are. In other words, if the different components of the inputs or the targets are correlated or of different variances, we expect GA to be hampered, observed in Sec.~\ref{sec:when}. Note that due to the $l-1$ exponent, we expect poor conditioning to have an even more drastic effect in deeper layers.

Notice that in this DRTP setup, the norm of the weights grows linearly with time, which makes DRTP inapplicable to regression tasks, and over-confident in classification tasks. It is clear in this case the the first layer learns the teacher, and the subsequent layers try to passively transmit the signal.

\section{Details about the experiments}

\subsection{Direct Feedback Alignment implementation}

We build on the Pytorch implementation of DFA implemented in~\cite{launay2020direct},
accessible
at~\url{https://github.com/lightonai/dfa-scales-to-modern-deep-learning/tree/master/TinyDFA}. Note
that we do not use the shared feedback matrix trick introduced in this work. We
sample the elements of the feedback matrix $F_l$ from a centered uniform distribution
of scale $1/\sqrt{n_l+1}$.

\subsection{Experiments on realistic datasets}
\label{app:degeneracy}

We trained 4-layer MLPs with 100 nodes per layer for 1000 epochs using vanilla SGD, with a batch size of 32 and a learning rate of $10^{-4}$. The datasets considered are MNIST and CIFAR10, and the activation functions are Tanh and ReLU. 

We initialise the networks using the standard Pytorch initialization scheme. We do not use any momentum, weight decay, dropout, batchnorm or any other bells and whistles. We downscale all images to $14\times 14$ pixels to speed up the experiments. Results are averaged over 10 runs.

\begin{figure*}[t!]
  \centering
    \begin{subfigure}[b]{\textwidth}
    \includegraphics[width=\linewidth]{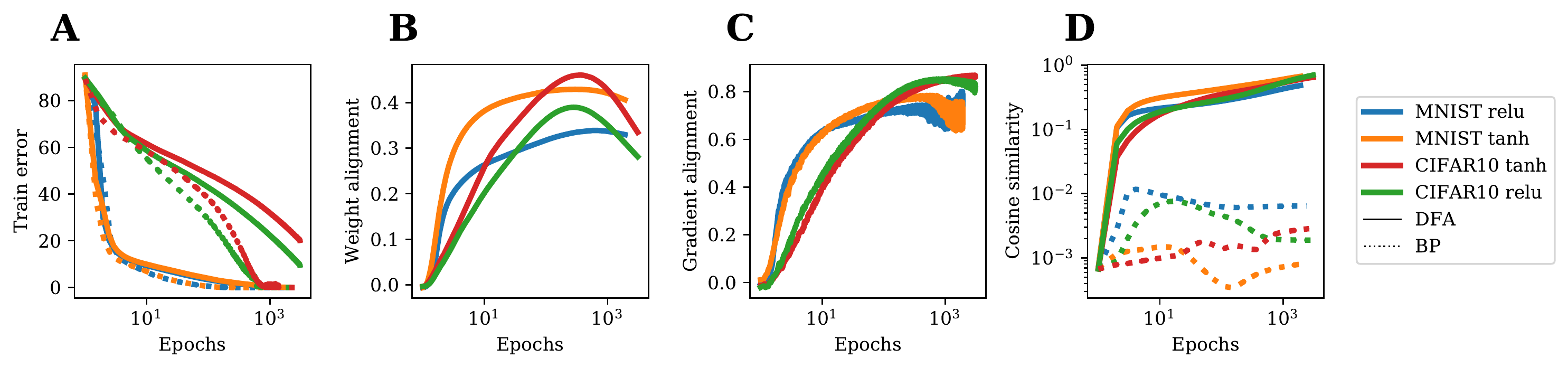}
    \caption{No label corruption}
    \end{subfigure}
\par\bigskip
    \begin{subfigure}[b]{\textwidth}
    \includegraphics[width=\linewidth]{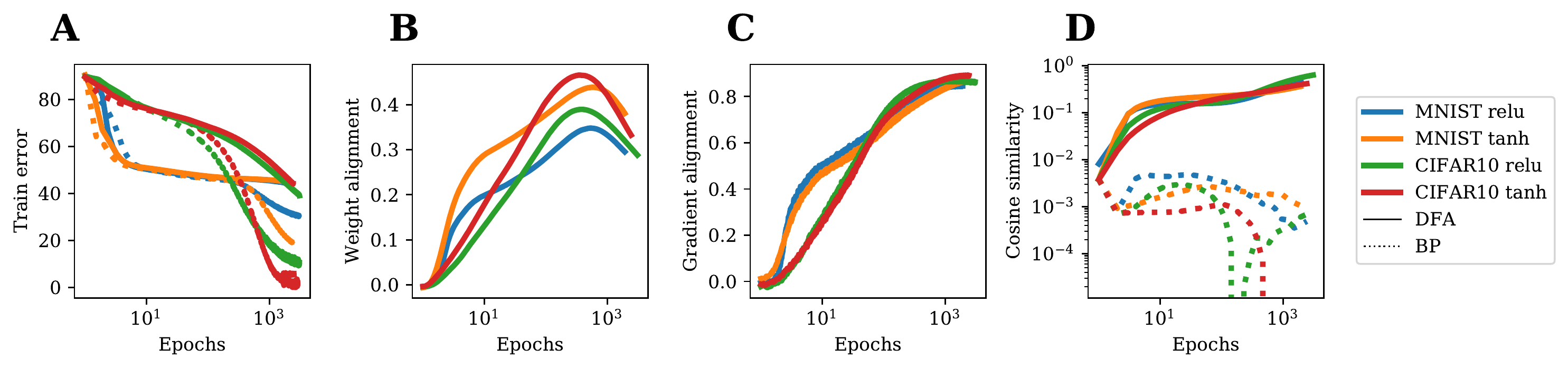}
    \caption{50\% label corruption}
    \end{subfigure} 
\par\bigskip
    \begin{subfigure}[b]{\textwidth}
    \includegraphics[width=\linewidth]{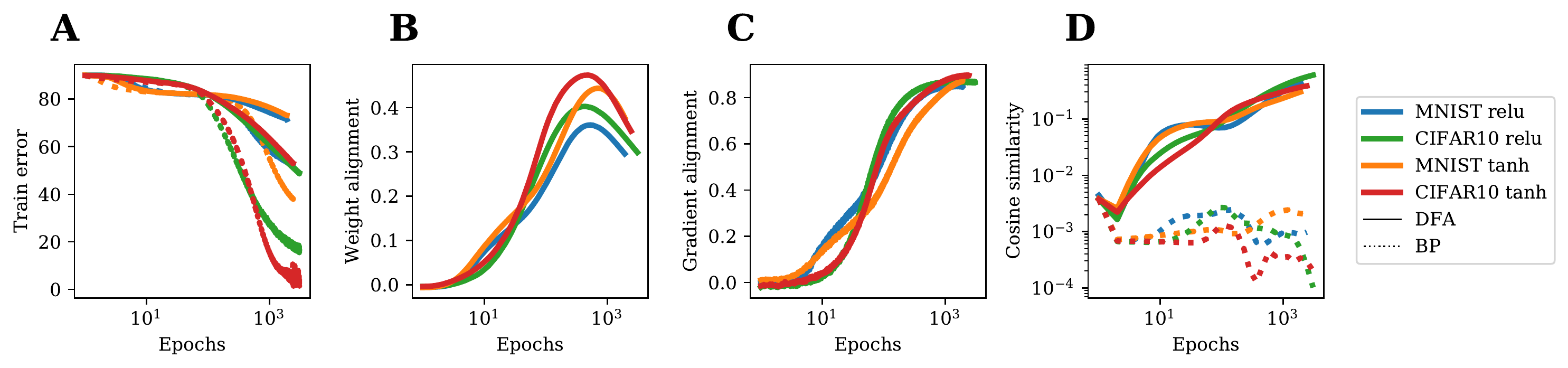}
    \caption{90\% label corruption}
    \end{subfigure} 
    \caption{Effect of label corruption on training observables. \textbf{A}:
      Training error. \textbf{B} and \textbf{C}: Weight and gradient alignment,
      as defined in the main text. \textbf{D}: Cosine similarity of the weight
      during training.}
\label{fig:dynamics-full}
\end{figure*}

For completeness, we show in Fig.~\ref{fig:dynamics-full} the results in the main text for 4 different levels of label corruption. The transition from Alignment phase to Memorisation phase can clearly be seen in all cases from the drop in weight alignment. Three important remarks can be made:
\begin{itemize}
    \item \textbf{Alignment phase}: Increasing label corruption slows down the early increase of weight alignment, as noted in Sec.~\ref{sec:data-dependence}. 
    \item \textbf{Memorization phase}: Increasing label corruption makes the datasets harder to fit. As a consequence, the network needs to give up more weight alignment in the memorization phase, as can be seen from the sharper drop in the weight alignment curves. 
    \item \textbf{Transition point}: the transition time between the Alignement and Memorization phases coincides with the time at which the training error starts to decrease sharply (particularly at high label corruption), and is hardly affected by the level of label corruption.
\end{itemize}

\subsection{Experiment on the structure of targets}
\label{app:targets}

We trained a 3-layer linear MLP of width 100 for 1000 epochs on the synthetic
dataset described in the main text, containing $10^4$ examples. We used the same hyperparameters as for the experiment on nonlinear networks. We
choose 5 values for $\alpha$ and $\beta$: 0.2, 0.4, 0.6, 0.8 and 1.

\begin{figure*}
    \centering
    \begin{subfigure}[b]{\textwidth}	    \includegraphics[width=.25\linewidth]{figures/wa/alphabeta_relu.pdf}
    \includegraphics[width=.7\linewidth]{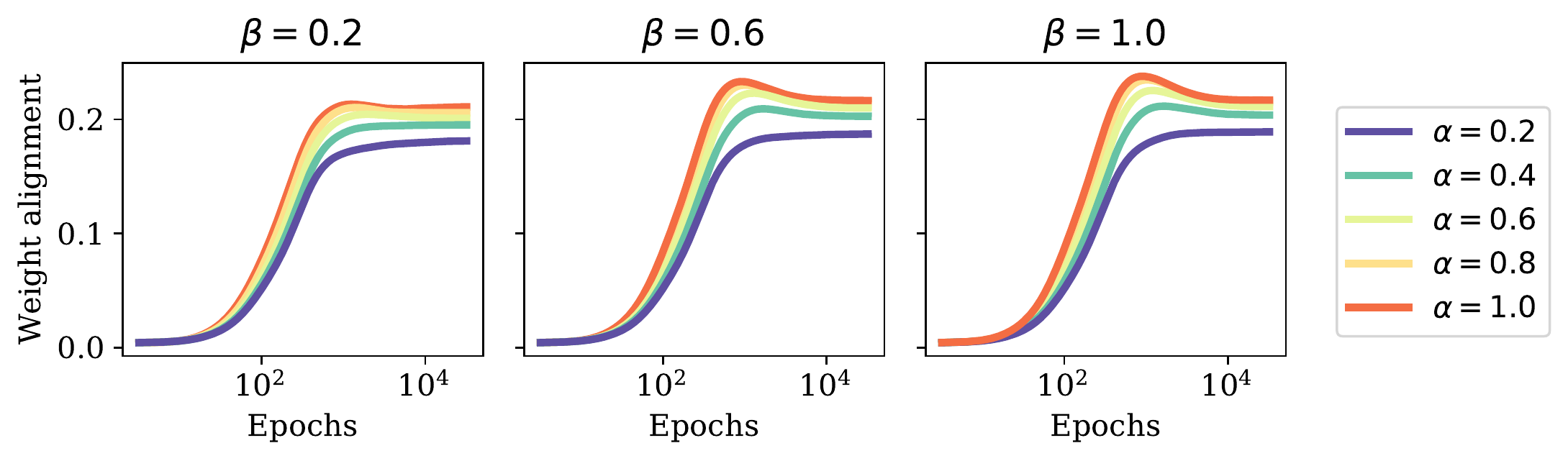}
    \caption{ReLU}
    \end{subfigure}  
    \begin{subfigure}[b]{\textwidth}	    \includegraphics[width=.25\linewidth]{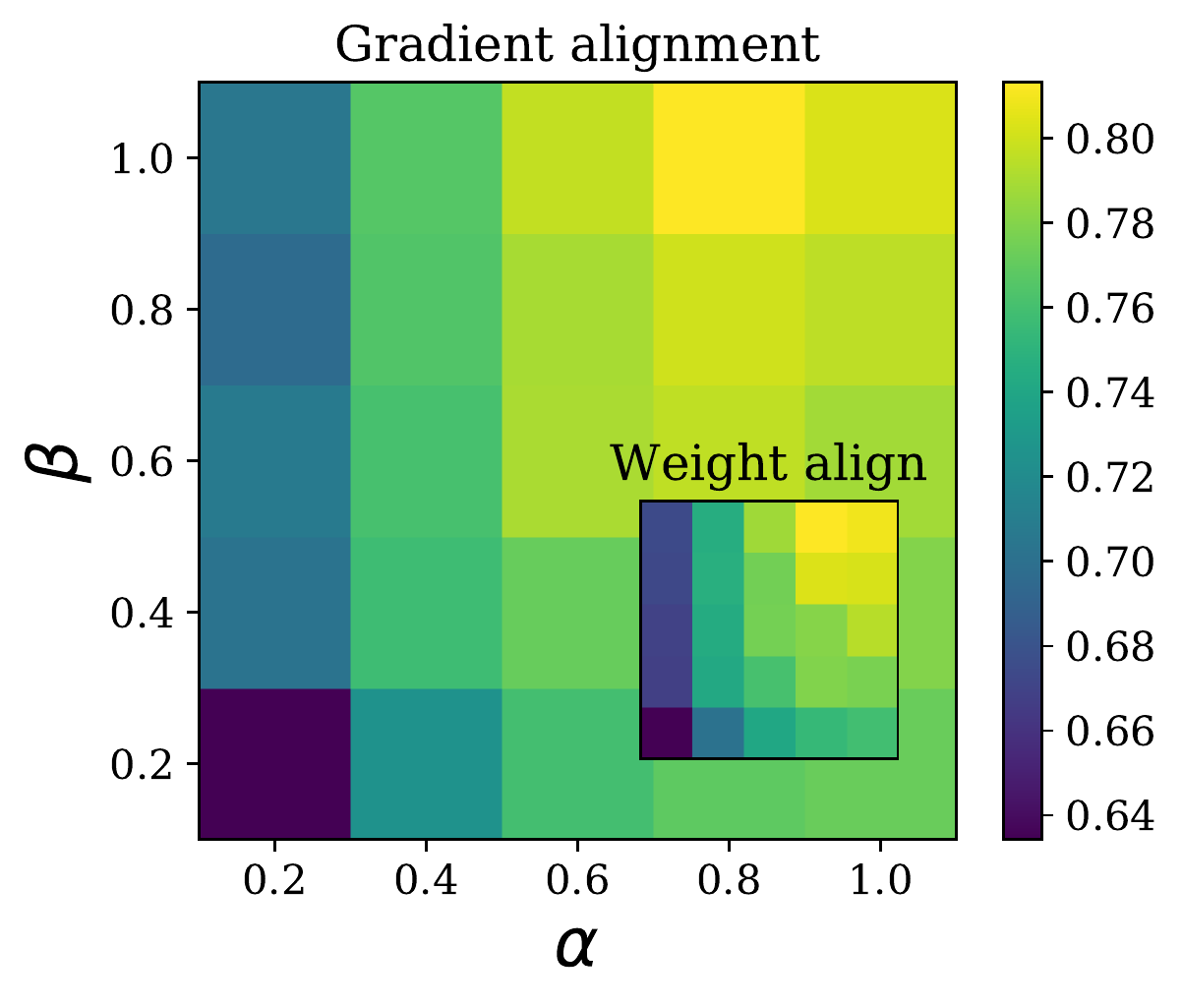}
    \includegraphics[width=.7\linewidth]{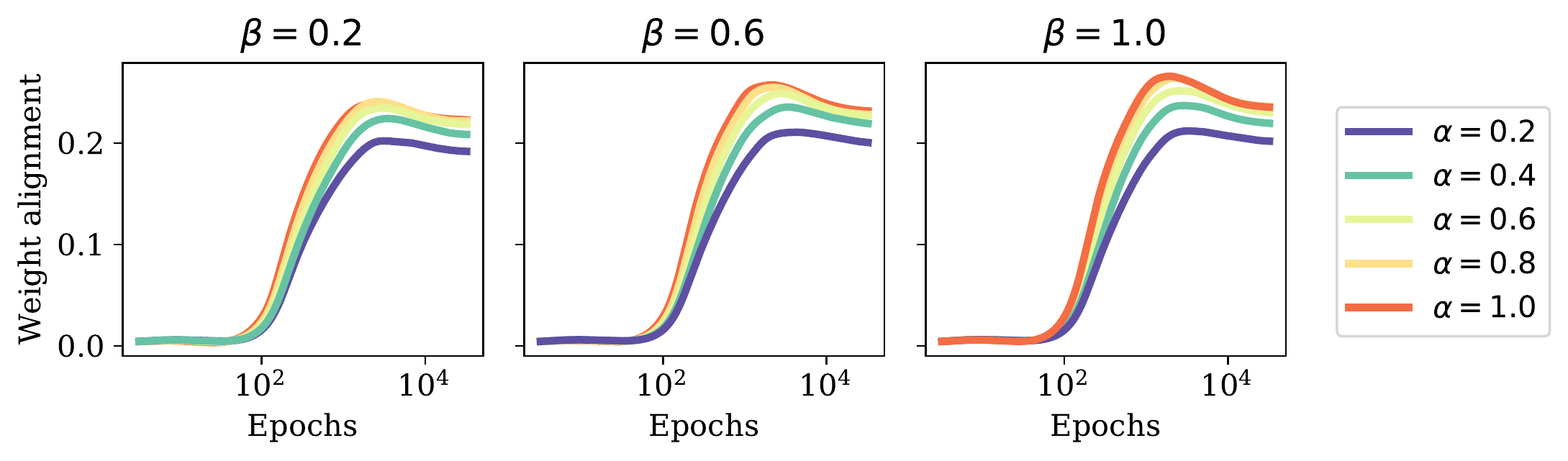}
    \caption{Tanh}
    \end{subfigure} 
\caption{WA is hampered when the output dimensions are correlated ($\beta<1$) or of different variances ($\alpha<1$).}
\label{fig:alphabeta-dynamics}
\end{figure*}

In Fig.~\ref{fig:alphabeta-dynamics}, we show the dynamics of weight alignment for both ReLU and Tanh activations. We again see the Align-then-Memorise process distinctly. Notice that decreasing $\alpha$ and $\beta$ hampers both the mamixmal weight alignment (at the end of the alignment phase) and the final weight alignment (at the end of the memorisation phase).

\end{document}